%% file: ICE_paper_v2.tex
\documentclass[11pt,a4paper]{article}

\usepackage[margin=1in]{geometry}

\usepackage{booktabs}
\usepackage{longtable}
\usepackage{placeins}
\usepackage{multirow}
\usepackage{amsmath}
\usepackage{amssymb}
\usepackage{graphicx}
\usepackage{xcolor}
\usepackage{enumitem}
\usepackage{microtype}
\usepackage{url}
\usepackage[htt]{hyphenat} 
\usepackage[round,authoryear,sort]{natbib}
\usepackage[colorlinks=true,linkcolor=blue!50!black,citecolor=blue!50!black,urlcolor=blue!50!black]{hyperref}
\usepackage{cleveref}
\usepackage{caption}
\usepackage{array}
\usepackage{ragged2e}
\usepackage{pgfplots}
\pgfplotsset{compat=1.18}
\usetikzlibrary{patterns,arrows.meta,positioning,fit,backgrounds,calc}

\tikzset{
  icebox/.style={draw=blue!40!black, fill=blue!5, rounded corners=2pt,
    align=center, font=\scriptsize, inner sep=3pt, minimum height=6mm},
  icephase/.style={draw=gray!50, fill=gray!8, rounded corners=3pt,
    align=center, font=\scriptsize\bfseries, inner sep=4pt},
  icestore/.style={draw=teal!50!black, fill=teal!8, rounded corners=2pt,
    align=center, font=\scriptsize, inner sep=3pt, minimum width=17mm, minimum height=9mm},
  iceleg/.style={draw=orange!60!black, fill=orange!8, rounded corners=2pt,
    align=center, font=\scriptsize, inner sep=2pt, minimum width=15mm, minimum height=6mm},
  icecore/.style={draw=red!50!black, fill=red!6, rounded corners=2pt,
    align=center, font=\scriptsize\bfseries, inner sep=3pt},
  iceflow/.style={-{Stealth[length=2mm]}, draw=gray!60, thick},
  iceflowd/.style={-{Stealth[length=2mm]}, draw=gray!60, thick, dashed},
}

\newcolumntype{L}[1]{>{\raggedright\arraybackslash}p{#1}}
\newcolumntype{C}[1]{>{\centering\arraybackslash}p{#1}}
\newcolumntype{R}[1]{>{\raggedleft\arraybackslash}p{#1}}

\usepackage{titlesec}
\titlespacing*{\section}{0pt}{1.4em}{0.7em}
\titlespacing*{\subsection}{0pt}{1.1em}{0.5em}

\renewcommand{\arraystretch}{1.15}
\title{\bfseries LSREP: A Longitudinal State-Replay Protocol for Evaluating Conversational Memory, with ICE v2 as an Audited Local-First Architecture}

\author{Deepesh Sonar\\[2pt]
  \normalsize Thakur College of Engineering and Technology, Computer Engineering, Mumbai, India\\
  \normalsize\href{mailto:18deepnar@gmail.com}{\texttt{18deepnar@gmail.com}}}

\date{}

\input{generated/ablation_macros.tex}
\begin{document}
\maketitle
\thispagestyle{empty}

\begin{abstract}
Conversational memory changes during use, so endpoint question answering alone cannot establish how a persistent state accumulates, ages, or incorporates revisions. We introduce \textbf{LSREP}, a Longitudinal State-Replay Evaluation Protocol combining ordered replay, explicit lifecycle schedules, repeated probes, evolving reference answers, and mechanism-fidelity checks. Its architectural case study is \textbf{ICE v2}, a local-first memory middleware with typed stores, retrieval fusion, and dynamic context budgets. The private, single-user instantiation contains 1,985 turns, 219 distinct probes, and 1,211 probe--checkpoint observations across 52 checkpoints. On three ordinary-density datasets, ICE v2 has a near-zero mean quality difference from vector-RAG while selecting 32\% fewer fragments but using 6.6\% more estimated prompt tokens. A fourth, dense dataset exposes catastrophic failures of the unbudgeted baseline. The fidelity audit limits attribution: procedural retrieval is defective, several mechanisms are unexercised, and graph utility is not established. In a complementary matched public diagnostic, ICE v2 loses decisively to pure vector-RAG on LongMemEval: 50.8\% versus 72.8\% in the evidence-only oracle and 43.0\% versus 69.5\% in full-S. Paired differences are $-22.0$ points (95\% CI $[-26.6,-17.4]$) and $-26.5$ ($[-31.3,-21.8]$). Conservative abstention accompanies severe multi-session and temporal failures. ICE uses less context in this diagnostic, establishing a quality--cost trade-off rather than superior efficiency. Together, replay, fidelity auditing, and public endpoint testing expose distinct failure modes that neither architectural descriptions nor aggregate scores identify alone.
\end{abstract}

\section{Introduction}
\label{sec:intro}

Absent an explicit persistence layer, a language model can use only what remains in its current context. Enlarging that context does not make every token equally usable: models attend unevenly across long inputs and degrade on information away from their edges~\citep{liu2024lost}, while effective context is often shorter than the advertised limit~\citep{hsieh2024ruler}. Commercial assistants now provide cross-session memory, so literal session-level amnesia is no longer a universal description. The open questions are instead what is retained, how revisions and forgetting are represented, and whether the user can inspect and govern the store.

Research has approached those questions through multi-session dialogue~\citep{xu2022msc,jang2023chronicles}, long-history benchmarks~\citep{maharana2024locomo,wu2025longmemeval}, and personalised adaptation~\citep{salemi2024lamp}. These studies examine complementary aspects of long-term interaction. Endpoint QA over supplied histories can test knowledge updates and temporal reasoning, but does not by itself require repeated observations of a reconstructed memory state under a declared maintenance and access schedule.

The Infinite Context Engine (ICE v2) is a local-first middleware around an otherwise stateless answering model. It stores episodic turns, a temporally-versioned knowledge graph, procedural patterns, documents, and user-controlled memory slots; classifies each prompt; retrieves and fuses candidates; and assembles a bounded context. Externalising memory preserves it when the answering model changes and makes the store inspectable. ICE is the case study here, not the definition of the evaluation problem.

That evaluation problem is longitudinal. A system may retrieve well at one endpoint while failing to update an old fact, may appear efficient only because a component never ran, or may work inside one continuing conversation but fail when a fresh session must aggregate earlier ones. We therefore introduce LSREP alongside, rather than in place of, fixed-history benchmarks. It reconstructs state at successive checkpoints and allows the reference answer itself to evolve.

This paper makes three contributions.

\begin{enumerate}[leftmargin=*]
  \item \textbf{LSREP}, the primary contribution: a system-independent protocol specification for replaying conversational history, reconstructing state at successive checkpoints, and scoring against evolving ground truth. It is designed for other systems and exported histories; empirical portability remains to be tested, and the private corpus used here is not independently reproducible.
  \item \textbf{ICE v2 as an architectural contribution and audited case study}: a local-first multi-store architecture with intent-driven retrieval, access-weighted decay, weighted rank fusion, and a per-query token budget. The evaluated snapshot is fixed at tag \texttt{v2-paper-eval}.
  \item \textbf{A two-regime evaluation with a fidelity audit}: LSREP measures within-conversation longitudinal behaviour and a density stress case; matched public LongMemEval oracle and full-S conditions probe fresh-session aggregation and distractors. Their disagreement is itself a result. A component audit separately identifies defective, unexercised, contributing, and inconclusive mechanisms.
\end{enumerate}

The methodological claim is deliberately falsifiable: a near-zero ablation delta supports ``does not help'' only when the component is verified live and the interaction regime reaches it. We report where our own evaluation failed that standard and constrain the system claims accordingly.

\paragraph{Research questions.}
\textbf{RQ1}: How do answer quality and context use change as ICE v2's replayed state accumulates and ages?
\textbf{RQ2}: Which mechanisms execute, affect the supplied evidence, and have an identifiable quality effect?
\textbf{RQ3}: Does the observed behaviour transfer to endpoint QA over separately supplied sessions, with and without distractors?
The third question concerns transfer between regimes, not a competition between evaluation protocols. All substantive system results concern frozen ICE v2; the v1 pilot is historical, and ongoing v3 development is outside this paper.

\section{Related Work}
\label{sec:related-work}
\subsection{Persistent and Structured Memory}
\label{sec:related-memory}
MemGPT manages a bounded context through a hierarchy of working and external memory~\citep{packer2023memgpt}. MemoryBank combines memory storage with time-dependent forgetting and reinforcement~\citep{zhong2024memorybank}. Generative Agents uses observations, reflection, and retrieval to support simulated agents~\citep{park2023generative}. These establish persistent memory and memory lifecycle management as existing ideas; ICE v2's contribution is their integration into an inspectable local middleware and the audit of that integration.

Mem0 explicitly extracts and updates memories, including ADD, UPDATE, DELETE, and NOOP operations; its graph variant adds relational representations, and its cited evaluation uses LoCoMo~\citep{chhikara2025mem0}. It should therefore not be characterised as append-only or unable to reconcile contradictions. Zep's Graphiti engine represents temporally qualified relationships and combines episodic, semantic, and community information~\citep{rasmussen2025zep}. Temporal graph memory is consequently not unique to ICE v2. The relevant architectural choices here are typed stores, explicit retrieval control, local deployment, and the observable boundary between an implemented mechanism and an evaluated capability.

\subsection{Graph and Adaptive Retrieval}
\label{sec:related-kg}\label{sec:related-rag}\label{sec:related-primitives}
GraphRAG uses extracted graphs and community summaries for query-focused summarisation~\citep{edge2024graphrag}; KGP navigates passage graphs for multi-document QA~\citep{wang2023kgp}; Think-on-Graph explores reasoning paths over a knowledge graph~\citep{sun2024tog}; HippoRAG combines a graph with personalised PageRank for associative retrieval~\citep{gutierrez2024hipporag}. These works motivate structural retrieval, but their reported experiments do not establish the correctness of ICE v2's evolving graph. That must be measured in the implementation under study.

RAG, REALM, Fusion-in-Decoder, and REPLUG develop different couplings between retrieval and generation~\citep{lewis2020rag,guu2020realm,izacard2021fid,shi2024replug}. CRAG evaluates retrieved evidence, and Self-RAG learns retrieval and critique decisions~\citep{yan2024crag,asai2024selfrag}. These methods are not intrinsically incompatible with changing corpora. LSREP adds a specification of when that corpus and the system's derived state change, and what answer is valid at each observation.

ICE v2 builds on lexical ranking, weighted reciprocal rank fusion (RRF), and an optional HyDE rewrite~\citep{robertson2009bm25,cormack2009rrf,gao2023hyde}. Its lexical leg uses PostgreSQL full-text ranking, historically named ``BM25'' in the code; it is not an implementation of the canonical BM25 scoring formula. RRF combines ranks without requiring comparable native scores. The v2 ablation supports a corrective effect after adding the lexical leg; HyDE was not isolated (Section~\ref{sec:ablation}).

\subsection{Evaluation Regimes}
\label{sec:related-eval}
Multi-Session Chat evaluates dialogue that uses earlier sessions~\citep{xu2022msc}; Conversation Chronicles incorporates time intervals and speaker relationships~\citep{jang2023chronicles}. LoCoMo evaluates long conversational histories through QA and other tasks~\citep{maharana2024locomo}; LongMemEval separately tests information extraction, updates, multi-session reasoning, temporal reasoning, and abstention~\citep{wu2025longmemeval}. LaMP studies personalisation from user profiles~\citep{salemi2024lamp}. Their targets differ, and they should not all be reduced to static single-shot retrieval.

LSREP's proposed unit is the \emph{state trajectory}: a specified history prefix, lifecycle schedule, reference version, and access history at each checkpoint. LongMemEval's supplied histories allow public endpoint comparisons that this paper's private replay corpus cannot provide. Conversely, repeated checkpoint observations allow us to inspect changes that a single endpoint score leaves unresolved. Neither regime subsumes the other.

LLM judging makes free-form evaluation practical but introduces position, verbosity, and other biases~\citep{zheng2023judging,liu2023geval,wang2024unfair}. Blinding and fixed rubrics mitigate some risks; they do not establish unbiased labels. This distinction matters when comparing our Muse-judged LongMemEval results with published GPT-4o-judged results.

\section{LSREP: Requirements, Algorithm, and Validity}
\label{sec:exp-design}\label{sec:lsrep}
\subsection{Evaluation Object and Requirements}
LSREP evaluates a \emph{reconstructed trajectory}, not an exact recovery of an unlogged historical deployment. Let $H_{\leq t}$ contain the supplied turns available by checkpoint $t$, $S_t$ the persistent state, $L_t$ the declared maintenance and ageing schedule, and $A_t$ the access events allowed to affect memory. A replay implementation realises
\[
S_t = F(S_{t-1}, H_{(t-1,t]}, L_t, A_{t-1};\theta),
\]
where $\theta$ fixes the system version, models, configuration, and runtime contract. A probe $q$ has an origin checkpoint $o(q)$ and a reference $G(q,t)$ derived only from $H_{\leq t}$. The observation is the answer, its score against $G(q,t)$, the actual context supplied, and the observable state transition caused by the query.

An admissible system must expose an ingestion path, a query interface, a reset or snapshot mechanism, and enough control to declare maintenance and query side effects. A chat endpoint alone is insufficient. The protocol requires ordered history, explicit checkpoint boundaries, temporally grounded references, consistent treatment of conditions, and versioned evidence of what ran. It does not require graphs, decay, or any ICE-specific store: a vector index is a valid instance whose lifecycle may simply append turns.

\subsection{Protocol Algorithm}
\begin{figure}[htbp]
\centering\fbox{\begin{minipage}{0.93\linewidth}\small
\textbf{Algorithm 1: Longitudinal state replay}\par
\textbf{Input:} ordered history $H$; checkpoints $T$; system conditions $C$; lifecycle schedule $L$; probes with origins; versioned references $G$; query-mutation policy $A$.\par
\begin{enumerate}[leftmargin=*,nosep]
\item Freeze code, model identities, parameters, clocks, and randomisation policy. Initialise each condition's declared state.
\item For each checkpoint $t\in T$, ingest only newly available turns, in order. Complete the declared write and maintenance jobs; record completion and failures.
\item Construct or load $G(q,t)$ for every eligible probe $o(q)\leq t$, using only the prefix $H_{\leq t}$. Preserve the probe's referent and evidence provenance when updating its answer.
\item Query all conditions under the same checkpoint and reference version. Record scope, selected context, costs, answer, and component activity. Apply the declared query side effects before the next observation, or restore a snapshot if side effects are excluded.
\item Judge answers with the fixed rubric and missing-output rule. Aggregate paired differences while retaining repeated-probe and conversation identities.
\item Report trajectories, quality--cost trade-offs, missingness, and mechanism-fidelity status. Separate unexercised mechanisms from executed failures and measured effects.
\end{enumerate}
\textbf{Output:} a versioned observation ledger and scoped claims about that replay process.
\end{minipage}}
\caption{LSREP specification. Assertions describe the validity contract; the retrospective ICE v2 audit identifies where the historical instantiation lacked sufficient instrumentation.}
\label{fig:lsrep-algorithm}
\end{figure}

Independent reconstruction of each checkpoint is also possible, but it must reproduce every permitted prior access event to represent the same trajectory. Replaying the same turns without those events is a different experimental condition. A deterministic schedule does not guarantee identical model outputs across executions.

\subsection{Evolving Ground Truth}
A reference records the tracked referent, supporting evidence available by the checkpoint, current answer, and prior versions. When a new turn revises a fact, a current-truth probe changes its reference; a question explicitly anchored to an earlier time retains its historical answer. Ambiguous revisions require adjudication or an uncertainty label, rather than silently changing the subject. References must remain outside the system's retrievable memory. Repeatedly answering an outdated reference correctly is not successful memory updating.

\subsection{Validity Conditions and Scope}
\label{sec:lsrep-standing}
\textbf{Temporal validity} prohibits future evidence in state or references. \textbf{Replay validity} requires an explicit lifecycle and access schedule, including whether evaluation queries reinforce memory. \textbf{Measurement validity} requires complete prompt accounting, inspectable missing-output rules, and judges that can discriminate relevant errors. \textbf{Mechanism validity} requires evidence that a claimed component executes and its output reaches a decision or answer context; non-empty output alone does not establish utility. \textbf{Statistical validity} requires stating the resampling unit and the population to which uncertainty applies.

The protocol was developed alongside ICE and instantiated on one user's private history. This creates selection and authoring risks even when the system obtains negative results: an unflattering result is not proof of an unbiased instrument. The v1 pilot exposed deficiencies in context accounting and judging (Appendix~\ref{app:exp1}); the v2 audit exposed defective and unexercised components. These motivate explicit validity conditions rather than establish universal validation of LSREP. Independent users, systems, and publicly releasable trajectories remain needed.

\section{Synthetic Worked Example}
\label{sec:worked-example}
The following example is invented and contains no private corpus text. It illustrates the protocol, not an additional ICE v2 experiment.
\begin{table}[htbp]
\centering\small
\caption{A reference changes only when the history available at that checkpoint warrants it.}
\label{tab:worked-example}
\begin{tabular}{@{}L{1.0cm}L{4.0cm}L{3.6cm}L{3.6cm}@{}}\toprule
Time & Newly available evidence & Current-truth probe/reference & Historical probe/reference\\\midrule
T1 & ``Project Atlas will use SQLite for storage.'' & ``Which database does Atlas use?'' $\to$ SQLite & ``What was the initial database choice?'' $\to$ SQLite\\
T2 & ``We have replaced SQLite with PostgreSQL for Atlas.'' & Same question $\to$ PostgreSQL & Same question $\to$ SQLite\\\bottomrule
\end{tabular}
\end{table}
At T1, answering PostgreSQL leaks future evidence. At T2, answering SQLite to the current-truth question repeats a superseded decision; answering PostgreSQL to the historical question erases the revision history. An answer that states both versions and their order may satisfy a richer evolution probe. The present v2 LSREP corpus tests current truth; historical and evolution probes in this illustration specify capabilities a future instantiation could test, not capabilities demonstrated by the present results.

\section{ICE v2 Architecture and Audit Contract}
\label{sec:architecture}

ICE v2 is an OpenAI-compatible memory middleware between a conversational client and a pool of locally-served models. It is not a model: every request addressed to the synthetic model name \texttt{ice-proxy} is intercepted by a proxy that classifies the turn, retrieves context from four long-lived memory stores, assembles a cache-friendly prompt, routes the request to a per-turn specialist model, streams the response back, and then dispatches background workers that extract, decay, cluster, and consolidate the new turn into long-term memory. The store may grow without the answering model's context window, but each request still receives a bounded, selectively rebuilt view of it. This section describes the mechanisms the evaluation exercises; Appendix~\ref{app:implementation} carries the implementation detail (classification engine, Codex internals, prompt assembly, worker cluster, operational infrastructure), and the archived technical report in the repository describes the system exactly as evaluated.

\begin{figure}[t]
\centering
\resizebox{0.92\columnwidth}{!}{%
\begin{tikzpicture}[node distance=5mm]
\node[icebox, minimum width=15mm, minimum height=9mm] (client) {Client\\request};
\node[icephase, right=8mm of client, minimum height=11mm, text width=62mm] (pre)
  {Pre-flight \textnormal{(synchronous, in request path)}\\[1pt]
   \textnormal{\scriptsize classify $\to$ override $\to$ retrieve $\to$ fuse (RRF)}\\
   \textnormal{\scriptsize $\to$ budget $\to$ assemble $\to$ route (MoE)}};
\node[icebox, fill=blue!12, right=8mm of pre, minimum height=9mm] (stream) {stream\\response};
\node[icephase, fill=orange!7, draw=orange!50, below=17mm of pre, minimum height=11mm, text width=62mm] (post)
  {Post-flight \textnormal{(asynchronous, after stream closes)}\\[1pt]
   \textnormal{\scriptsize store raw turn $\to$ evaluate representation}\\
   \textnormal{\scriptsize $\to$ summaries and typed-memory extraction}};
\node[icestore, minimum width=22mm, minimum height=11mm] (db) at (stream |- post) {PostgreSQL\\+ pgvector};
\draw[iceflow] (client) -- (pre);
\draw[iceflow] (pre) -- (stream);
\draw[iceflowd] (stream.south) -- ++(0,-7mm) -| (post.north) node[pos=0.25, above, font=\scriptsize\itshape, gray] {dispatch};
\draw[iceflow] (post) -- (db);
\draw[iceflowd] (db.east) -- ++(10mm,0) |- ([yshift=7mm]pre.north) -- (pre.north) node[pos=0.70, above, font=\scriptsize\itshape, gray] {read state};
\end{tikzpicture}%
}
\caption{System overview. Pre-flight runs synchronously in the request path; post-flight runs asynchronously after the stream closes and writes memory for later retrieval. Scheduled maintenance ages and consolidates that state.}
\label{fig:system-overview}
\end{figure}
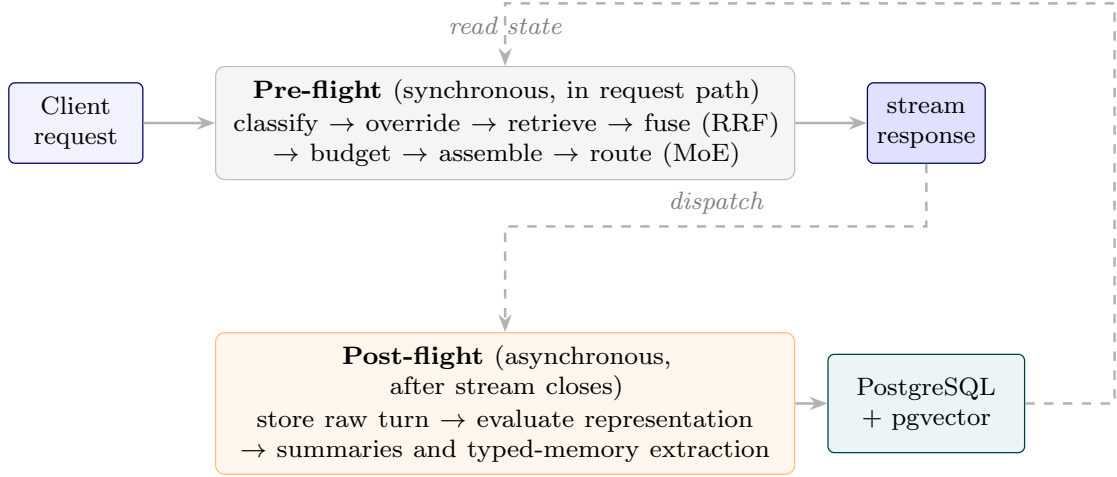

\subsection{Request Lifecycle}

Each turn traverses a \textbf{pre-flight} phase (synchronous, in the request path) and a \textbf{post-flight} phase (asynchronous, after the stream closes).

In pre-flight, the user message is classified into topic tags, intent tags, and a single context-reliance label (Zero\_Shot, Long\_Term\_Memory, Real\_Time\_Search) by a two-stage cascade: a rule-based pre-classifier resolves rule-matching cases, and on a miss a small multi-layer perceptron over a frozen \texttt{Qwen3-Embedding-0.6B} encoder resolves the remaining cases (encoder time must be accounted for separately). Hard override rules then coerce the label---most consequentially, a conservative rule upgrading Zero\_Shot to Long\_Term\_Memory when a conversation exceeds ten turns or confidence falls below 0.95, so retrieval runs for almost every turn in a long conversation; Section~\ref{sec:limitations} discusses what that costs. A Hybrid Retrieval Orchestrator then runs the retrieval legs, fuses them with weighted Reciprocal Rank Fusion, and post-processes the fused list. A Prompt Assembler concatenates the surviving fragments with persistent memory slots, recent turns, and the live message under a stable prefix, and a Mixture-of-Experts router selects a locally-served model with per-conversation stickiness.

In post-flight, an evaluator runs lossless detection---the guiding principle in the code is that \emph{memory is earned}: raw turn text is stored first; density and other rules govern whether raw text or a summary is later injected---and dispatches idempotent extraction tasks for the knowledge graph and for procedural patterns. Scheduled workers (decay, clustering, reflection, batch summarisation, sentinel monitoring, and a weekly classifier fine-tune) maintain the stores over time; \cref{tab:workers} in the appendix lists them.

\subsection{Memory Stores}

ICE maintains four stores plus two structured overlays, all in one PostgreSQL database with the pgvector extension; every vector column is 384-dimensional because a single embedder is shared throughout.

\begin{table}[htbp]
\centering
\small
\caption{The four memory stores and the overlays. Each store is queried by one or more retrieval legs. Appendix~\ref{app:implementation} gives schemas and write paths.}
\label{tab:stores}
\begin{tabular}{@{}L{2.2cm}L{6.0cm}L{4.4cm}@{}}
\toprule
\textbf{Store} & \textbf{Purpose} & \textbf{Retrieval method} \\
\midrule
Episodic & Every conversational turn: raw text, summary, lossless flag, decay score, access count, embedding & BM25 leg; vector leg with decay weighting; session diversification \\
Codex & Temporally-versioned semantic graph: entities with aliases and properties, typed edges carrying \texttt{valid\_from}/\texttt{valid\_until}, append-only event log, snapshots & Graph traversal (BFS depth 3 over edges where \texttt{valid\_until IS NULL}); enumeration fallback \\
Procedural & Recurring behavioural patterns: trigger conditions, reinforcement count, confidence & Vector top-5 behind a hard intent gate \\
RAG & Externally-ingested documents, chunked and embedded & Vector top-5, triple-gated \\
\midrule
Memory slots & Seven server-enforced persistent slots (persona, preferences, project context, guidance, pending items, \ldots) & Prepended verbatim to every prompt \\
Context clusters & Conversation-scoped topical groupings with unit-norm centroids & Restrict episodic search to relevant clusters \\
\bottomrule
\end{tabular}
\end{table}

The \textbf{Codex} is the store most specific to conversation. Its central design lever is a three-bucket controlled relation vocabulary---\emph{property} relations overwrite, \emph{multi-valued} relations accumulate, and \emph{single-valued} relations auto-expire the prior edge of the same (source, relation) pair---with generic relations such as \texttt{is} and \texttt{has} deliberately absent so the extractor is forced to be specific. Setting \texttt{valid\_until} rather than deleting is what makes the graph temporal: an edge with \texttt{valid\_until IS NULL} is treated by the store as current (not independently verified true), and a non-NULL value means historically superseded but still auditable. Appendix~\ref{app:codex} details the write path and entity resolution.

\textbf{Decay} is access-weighted. Three decay workers apply multipliers corresponding to roughly 5\%, 2\%, and 1\% decay per day for unaccessed, accessed, and creative-tagged turns respectively, with a \emph{creative floor} clamping decay score at 0.3 for narrative turns regardless of age---long-form fiction should not be summarised away for being old. Turns below 0.1 are archived and below 0.05 moved to cold storage. Codex edges decay similarly and are demoted from active to pending below a strength threshold; procedural patterns deactivate after 180 days with fewer than three reinforcements. Crucially, decay is not one-directional: every retrieval increments a fragment's access count and restores 0.15 to its decay score, so \textbf{retrieval is a partial reversal of forgetting} and what the system used yesterday is cheaper to reach today.

\subsection{Retrieval, Fusion, and Budget}
\label{sec:orchestrator}
The v2 design has six retrieval legs: lexical and decay-weighted vector episodic retrieval, graph traversal, procedural lookup, document lookup, and batch summaries. ``BM25'' below is the historical name of the PostgreSQL lexical leg. The two episodic legs are operational; procedural lookup is defective, and document and batch-summary stores were empty in LSREP. Graph traversal returns content, but its quality contribution is inconclusive. The full design must not be read as six validated mechanisms.

For candidate $f$, weighted fusion computes $s(f)=\sum_\ell\alpha_\ell/(60+\operatorname{rank}_\ell(f))$, with intent- and topic-dependent weights. Fragment text hashes identify duplicates. Post-fusion bonuses, session diversification, and greedy packing determine the selected set. The diversity pass operates on \emph{source type}: both episodic legs share one type, so it does not guarantee one item from every retrieval leg.

\begin{figure}[htbp]
\centering\fbox{\begin{minipage}{0.93\linewidth}\small
\textbf{Algorithm 2: ICE v2 context selection}\par
\begin{enumerate}[leftmargin=*,nosep]
\item Classify the query with current-conversation context; apply retrieval overrides. Derive the recent and retrieved-context budgets from that conversation's turn count, density, and labels.
\item If the low-confidence fallback fires, use its wide-net path and 2,000-token ceiling. Otherwise obtain scoped candidates from enabled legs and combine their rankings by weighted RRF.
\item Apply bonuses, cap each foreign conversation at three fragments, and deduplicate text. The active conversation is uncapped by this session rule.
\item Admit the best candidate per source type when it fits; greedily fill the remaining retrieval budget. Strengthen selected episodic memories.
\item Assemble system instructions, active slots, a bounded recent window, selected fragments, boundary acknowledgement, and the live query.
\end{enumerate}
\end{minipage}}
\caption{Frozen ICE v2 selection contract. Candidate generation and selection occur before answer generation; execution is not evidence of answer-quality benefit.}
\label{fig:retrieval-algorithm}
\end{figure}

The nominal context allowance is 23,000 estimated tokens with a 1,800-token overhead reserve. A growth cap limits retrieval to 2,000 estimated tokens for an empty current conversation, regardless of how much history exists in other conversations. This makes the transition from continuing-use LSREP to fresh-session LongMemEval architecturally consequential. Unused allowance is not reallocated. Token estimates use $\lfloor1.33\times\text{word count}\rfloor$; they are not model-tokenizer limits.

\subsection{Local Operation and Audit Contract}
The v2 deployment uses FastAPI, PostgreSQL/pgvector, local model serving, and Celery/Redis workers. The cloud answerer and judge used in the later public diagnostic are evaluation substitutions; they do not make that diagnostic an all-local run. The stores and control interfaces make state inspectable, while lossless flags, graph validity intervals, retrieval traces, and configuration pins provide audit surfaces.

A fidelity claim requires tracing input eligibility, execution, produced output, selection, and final prompt inclusion. A quality attribution additionally requires a controlled contrast in which the mechanism actually changes. A failed leg, an empty store, a constant ablation flag, and an executed component with an uncertain quality effect are separate outcomes. The historical harness did not enforce this full contract; RQ2 reports the retrospective audit rather than retroactively claiming it did.

\section{Evaluation Instantiation}
\subsection{Four LSREP Datasets}
\label{sec:datasets}

LSREP uses four long-form conversational datasets authored by one user during sustained real work. A turn is one user prompt paired with its assistant response. Internal conversation names and identifiers are replaced by labels A--D; the manuscript includes no verbatim private turn or probe text. \cref{tab:datasets} reports the corpus size, probe--checkpoint observation count, checkpoint count, and simulated decay horizon used by the harness.

\begin{table}[htbp]
\centering
\small
\caption{ICE v2 LSREP datasets and evaluation coverage. Observations repeat 219 distinct probes (A: 45; B: 91; C: 37; D: 46). Token counts use the harness's documented estimate, $1.33\times$ whitespace-delimited words, rather than a model-specific tokenizer. ``Days'' is the maximum simulated decay horizon, not elapsed wall-clock time.}
\label{tab:datasets}
\resizebox{\textwidth}{!}{%
\begin{tabular}{@{}lL{4.0cm}rrrrr@{}}
\toprule
\textbf{Label} & \textbf{Description / principal stress} & \textbf{Turns} & \textbf{Est. tokens} & \textbf{Obs.} & \textbf{Ckpts.} & \textbf{Days} \\
\midrule
A & Creative writing; narrative continuity and character evolution & 290 & 237K & 173 & 10 & 24 \\
B & Long-form world-building; horizon retrieval and entity evolution & 1{,}119 & 806K & 638 & 20 & 93 \\
C & Technical planning; information density and 8K+ token turns & 325 & 785K & 154 & 10 & 27 \\
D & Academic planning; personal/technical decisions & 251 & 155K & 246 & 12 & 20 \\
\midrule
\textbf{Total} & & \textbf{1{,}985} & \textbf{1.98M} & \textbf{1{,}211} & \textbf{52} & --- \\
\bottomrule
\end{tabular}%
}
\end{table}

The four were selected before scoring for three properties: 251--1{,}119 turns, so relevant material can be separated by hundreds of exchanges; complementary domains; and recurring entities, revisions, procedures, and long-range dependencies. The replay normalises timestamps and applies the simulated horizons in the final column; these are experimental ageing schedules, not claims about the original conversations' calendar duration. Aggregate reports and harness code are released, while raw conversations and probe text remain private because they contain the sole author's personal data. LongMemEval supplies the complementary public corpus in Section~\ref{sec:lme-oracle}.

\subsection{ICE v2 Replay and Answer Conditions}
\label{sec:exp2-env}

\textbf{System under test.} Experiment 2 evaluated frozen ICE v2: the classifier backbone upgraded to Qwen3-Embedding-0.6B with the head retrained from scratch; a trained micro-NER model replacing the pilot's regex entity tagger, so the graph extractor grounds on real entity detection (loaded checkpoint; no task-specific NER recall estimate is established here); the conservative routing override added; the static budget and uncapped sliding window replaced by the unified dynamic token budget of Section~\ref{sec:orchestrator}; and cluster-scoped retrieval introduced.

\textbf{Protocol instantiation.} Rather than many conversations at isolated checkpoints, Experiment 2 follows four long-horizon conversations across their entire lifespan. Three choices differ sharply from the pilot. \emph{(1)~Continuous replay with preserved state}: each conversation is replayed chronologically into one fresh deployment, and memory stored at an early checkpoint remains available at every later one; a simulated memory lifecycle---decay, edge decay, procedural decay, reflection, clustering, consolidation, sentinel monitoring---runs \emph{between} checkpoints, so retrieval reflects aged, consolidated memory rather than fresh storage. \emph{(2)~Automatic, leg-aware probe generation}: probes are generated at each checkpoint from a sampled history window (first 5 turns, last 10 before the checkpoint, 15 random from between) using only information available up to that checkpoint, explicitly targeting specific legs, and each requires a unique temporal anchor to prevent ambiguity across hundreds of turns. These are combined with 72 hand-written probes inherited from the pilot. \emph{(3)~Incremental evaluation}: every probe whose origin is at or before a checkpoint is re-evaluated at that checkpoint, so a probe generated early is scored repeatedly as history accumulates---which is what makes longitudinal memory growth directly observable.

\textbf{Evolving ground truth.} The pilot's single frozen reference is replaced by a three-stage temporal-refinement pipeline: \emph{forensic regeneration} of the origin-checkpoint answer under a stricter, evidence-oriented prompt that records temporal provenance; \emph{temporal anchoring}, attaching to each probe an anchor identifying the fact or event it tracks; and \emph{forward propagation}, updating each answer at every later checkpoint when new content contradicts, updates, or extends it, with anchor-preservation checks guarding against silent drift to a superficially similar later event. Each probe therefore carries a temporally evolving reference document rather than a static answer---necessary because otherwise a system that correctly \emph{updated} a fact would score worse than one repeating an outdated one.

\textbf{Scoring.} Experiment 2 adds temporal-aware scoring (an answer loses points for reporting a superseded fact even if it was once correct) and inference credit (a specific, evidence-consistent detail not explicitly in the reference is rewarded rather than flagged as fabrication).

\textbf{Human scoring and label audit.} The released aggregation code merges 72 manual evaluation records, replacing their absolute scores and tournament rankings where supplied. These records therefore contribute to the headline aggregates; the aggregate is not purely automated. Separately, the reported v2 hallucination labels were reviewed against source conversations to correct false positives. The hand-scored subset is not an independent population or a second blinded annotator study. No corresponding manual correction was applied to the ablation's hallucination labels.

\textbf{Judge.} Scoring free-form answers with a language model follows now-standard practice~\citep{zheng2023judging,liu2023geval}, and inherits its known failure modes: judges are sensitive to answer position and verbosity and are not fair evaluators by default~\citep{wang2024unfair}. We mitigate rather than assume this away---anonymised four-condition tournaments, a fixed rubric at temperature 0.0, a judge model disjoint from every answering model, and the hand-audited slice described above; Section~\ref{sec:limitations} states what remains uncontrolled. The LSREP generalist answerer is \texttt{gemma4:26b-a4b-it-q4\_K\_M}. Both LSREP experiments use the same independent judge, \texttt{gemma-4-12B-AWQ} served on SGLang with a 150{,}000-token context at temperature 0.0 (requested deterministic decoding; execution-level reproducibility is not established), never used for retrieval, generation, probe generation, or ground-truth construction. The extraction rules---strict neutrality, verbatim anchor preservation, temporal dominance, third-person attribution---are shared; Experiment 2 applies them through the evolving-dossier pipeline rather than a single pass.

\textbf{Conditions.} A four-condition matrix: \textbf{Vector RAG (generalist)}---single-leg pgvector similarity, top-30, no decay weighting, classification, or fusion, on a generalist model; \textbf{Vector RAG (MoE)}---same retrieval, expert-routed; \textbf{Full ICE (generalist)}---all legs enabled, RRF fusion, classification, decay weighting, dynamic budget, post-fusion curation; and \textbf{Full ICE (MoE)}. Both use the same embedder and database. The baseline is a standard single-leg vector-RAG---top-30 similar turns assembled with the question---while the ICE conditions run the full system, fused retrieval \emph{plus} ICE's prompt assembler. The comparison is therefore \emph{full memory system versus standard vector-RAG}, which is what a deployment decision actually looks like: ICE's curation deliberately governs what context is assembled, not merely which fragments are retrieved.

\textbf{A note on replay.} Because history is \emph{replayed} rather than lived, the reinforcement that normally strengthens frequently referenced memories during use is largely absent---the primary reinforcement signal becomes the evaluation probes themselves. A concept referenced dozens of times in real use and one mentioned once therefore look more alike under replay than they would in deployment. This changes the state trajectory, but its direction is not guaranteed: reinforcement can promote useful evidence or amplify an early retrieval error. Experiment~2 should therefore be read as performance under the specified replay process, not as a lower or upper bound on deployment.

\subsection{Metrics and Statistical Units}
Scores use the historical 1--5 rubric. Win rate is the share of anonymised four-condition tournaments ranked first, not a direct two-arm win probability. Prompt tokens include the recent window and instruction scaffolding; fragment counts measure selected pieces, whose lengths differ. SPF (mean score divided by mean fragments) and TUR (mean score per thousand estimated prompt tokens) are descriptive ratios of an ordinal score, not standalone proofs of efficiency. We report quality and context use jointly.

The historical v2 LSREP confidence intervals use 10,000 paired bootstrap resamples of \emph{probe--checkpoint records}. Repeated observations of a probe are dependent, and conversations share one author; these intervals do not quantify generalisation across users or independent trajectories. Missing scores follow the archived chain: available score; otherwise failed answer $\to1$; otherwise sibling routing-condition score; otherwise rounded within-record mean; otherwise 3. The reported near-zero difference is absence of a detected mean-score difference, not a formal equivalence test.

LongMemEval uses binary correctness. Its intervals use 20,000 question-level paired percentile-bootstrap resamples, seed 20260911. Every resample retains both arms; the phase-change contrast retains all four arm--phase outcomes. Missing judgements are excluded from paired contrasts and bounded over all questions. Category intervals are exploratory and unadjusted for multiplicity. They condition on the recorded generations and judge verdicts; they do not include model rerun or judge-calibration uncertainty.

\subsection{Matched LongMemEval Setup}
\label{sec:lme-setup}
We use all 500 oracle questions and all 500 full-S questions~\citep{wu2025longmemeval}. Oracle supplies evidence sessions; full-S adds the benchmark's remaining history. Adapter \texttt{ice-v2-lme-sessions-v2} wipes the store per question, ingests each session as a separate auto-scoped conversation, and asks from a fresh empty conversation. Both arms read the same stored vectors. The pure vector-RAG arm retrieves the top 30 turn representations without decay weighting, fusion, or a token budget; it is a strong high-context comparison, not a matched-budget arm.

Both arms use OpenCode Go \texttt{gpt-5.6-luna} through the Responses endpoint, with a 4,096-output-token cap. The provider does not support the requested temperature parameter, so we do not claim temperature-controlled generation. ICE memory construction uses the evaluated Ollama \texttt{qwen3:4b-instruct-bg}. The judge is \texttt{muse-spark-1.3-contributor}, using the official LongMemEval task-specific judgement prompts and yes/no rule, with a 4,096-token judge cap. Muse is not the benchmark's official GPT-4o judge: these are within-study comparisons, not leaderboard-comparable scores. Judge identity, the decision prompt, and the answerer are distinct parts of the experimental stack.

Run-enablement changes restore the tag's background route to shared Ollama and run the same 384-dimensional embedder on CUDA (recorded CPU--CUDA cosine agreement 0.99986). The adapter calls v2 classification, budget setting, retrieval, and prompt assembly directly; it does not replay the entire HTTP lifecycle. In particular, it omits the API wrapper's faulty secondary word check (Appendix~\ref{app:assembly}). The measurement is of this documented v2 adapter path. No v3 repair is included.

The initial flattened-session adapter was invalidated after an identifier-type mismatch incorrectly capped the active conversation at three fragments. The admitted adapter preserves session boundaries and string identifiers, validates clean-store layout, and records its version. A historical local-Gemma oracle preceded the matched cloud run; the historical decision to stop before full-S was superseded when the matched study completed both phases. Appendix~\ref{app:lme-history} preserves that history without mixing its scores into RQ3.

\section{Results}
\label{sec:results}

\subsection{RQ1: Longitudinal Behaviour under LSREP}
\label{sec:exp2}

\begin{table}[htbp]
\centering
\small
\caption{ICE v2 results for all four datasets (generalist conditions). Scores are 1--5; tokens are estimated complete-prompt counts; wins are first places in four-condition tournaments. Dataset C is the density stress case.}
\label{tab:exp2-perconv}
\resizebox{\textwidth}{!}{%
\begin{tabular}{@{}lrrrrrr@{}}
\toprule
\textbf{Dataset} & \textbf{ICE score} & \textbf{Vec.\ score} & \textbf{ICE tok.} & \textbf{Vec.\ tok.} & \textbf{ICE win \%} & \textbf{Vec.\ win \%} \\
\midrule
A (Creative Writing)   & 3.63 & 3.50 & 19{,}434 & 24{,}360 & \textbf{37.2} & 18.0 \\
B (Long-Form Creative) & 4.28 & 4.33 & 24{,}108 & 21{,}257 & 32.8 & 22.9 \\
C (Technical Planning) & 4.33 & 1.23 & 19{,}953 & 88{,}061 & 43.5 & 1.9 \\
D (Academic Planning)  & \textbf{4.63} & 4.57 & 20{,}103 & 18{,}076 & 20.4 & 18.8 \\
\bottomrule
\end{tabular}%
}
\end{table}

\cref{tab:exp2-global} reports the four conditions on the three-conversation view (Dataset C excluded; 1{,}057 probe--checkpoint observations). Per-conversation, score-distribution, and temporal-quality breakdowns are in Appendix~\ref{app:exp2-detail}.

\begin{table}[htbp]
\centering
\small
\caption{Experiment 2, global comparison, Dataset C excluded (1{,}057 probe--checkpoint observations across 3 conversations). On the conversations where both systems can generate usable answers, ICE v2 has similar mean scores to the vector baseline, injects 32\% fewer fragments, and ranks first in 30.6\% of four-condition tournaments versus 21.2\%.}
\label{tab:exp2-global}
\resizebox{\textwidth}{!}{%
\begin{tabular}{@{}lccccccc@{}}
\toprule
\textbf{Condition} & \textbf{Score} & \textbf{Tokens} & \textbf{Frags} & \textbf{SPF} & \textbf{TUR} & \textbf{Win \%} & \textbf{Hall. \%} \\
\midrule
Vector RAG generalist & 4.25 $\pm$ 1.09 & 21{,}025 & 30.0 & 0.14 & 0.20 & 21.2 & 20.5 \\
Vector RAG MoE        & 4.24 $\pm$ 1.04 & 21{,}025 & 30.0 & 0.14 & 0.20 & 19.2 & 17.4 \\
Full ICE generalist   & 4.26 $\pm$ 1.08 & 22{,}411 & 20.4 & 0.21 & 0.19 & \textbf{30.6} & 19.6 \\
Full ICE MoE          & \textbf{4.28} $\pm$ 0.99 & 22{,}411 & 20.4 & 0.21 & 0.19 & 29.0 & 19.9 \\
\bottomrule
\end{tabular}%
}
\end{table}

The rounded mean scores are 4.26 and 4.25. The historical paired aggregation gives $\Delta=+0.002$, reported as $+0.00$ with a record-level 95\% CI $[-0.07,+0.07]$. The unrounded means are 4.25544 and 4.25355; rounding them separately makes their displayed difference 0.01. This is no detected mean-score difference under the archived analysis; repeated-probe dependence and manual-score merging limit its interpretation (Section~\ref{sec:exp2-env}).

\textbf{Repeated-probe sensitivity.} Resampling entire probe trajectories gives an ordinary-density paired interval of $[-0.148,+0.158]$ around $+0.002$ (182 probes), wider than the historical record-bootstrap interval. All-data and density-only clustered intervals remain positive: $[+0.192,+0.618]$ and $[+2.721,+3.470]$. These retain the qualitative findings but do not estimate variation across users (Appendix~\ref{app:clustered}).

\textbf{Scoring and ordinal sensitivity.} Removing the manual replacements gives an ordinary-density mean difference of $-0.020$ (probe-cluster 95\% CI $[-0.169,+0.136]$), consistent with no detected advantage. An ordinal comparison avoids assuming equal distances between rubric levels: across the 1,057 ordinary-density paired observations, ICE v2 scores higher on 216, vector on 215, and 626 tie. The net proportion favouring ICE is $+0.1$ percentage points (cluster CI $[-7.6,+8.0]$). In the archived merged scores, 130 vector-generalist failed answers receive score 1 and two scores use the sibling routing condition; ICE-generalist has all 1,211 explicit scores. Excluding every pair without two explicit scores reduces the all-data difference to $+0.046$ ($[-0.101,+0.199]$, $n=1,079$). This selected subset omits most density failures; it cannot replace the reliability analysis. Appendix~\ref{app:scoring} reports both views.

\textbf{Context use.} ICE v2 selects 20.4 rather than 30.0 fragments on average (32.0\% fewer), but uses 22,411 rather than 21,025 estimated prompt tokens (6.6\% more). Its recent window, persistent slots, and instruction structure are part of the full-system condition. The observed benefit is fragment economy at similar mean scores, not a token saving. A matched-prompt or matched-budget intervention was not run, so the costs and quality contributions of individual assembly choices remain unidentified.

\textbf{Tournament preference.} ICE v2 generalist is ranked first in 30.6\% of four-condition tournament appearances, versus 21.2\% for vector generalist. The historical record-bootstrap intervals are $[27.9,33.4]$ and $[18.8,23.7]$. This is not a direct head-to-head win rate, and shuffled presentation does not eliminate judge bias. Audited hallucination rates are 19.6\% and 20.5\%, respectively.

\textbf{Fragment--score association.} Recorded correlations are $r=0.193$ for ICE v2 generalist and $r=-0.015$ for vector generalist. The latter is near zero. These observational associations do not establish that curation removes noise, that longer answers are better, or that additional context causes higher quality.

\begin{figure}[t]
\centering
\begin{tikzpicture}
\begin{axis}[
    width=0.98\columnwidth, height=5.6cm,
    xlabel={\small Turn-index bin}, ylabel={\small Mean probe score},
    ylabel near ticks,
    xmin=0, xmax=1100, ymin=3.6, ymax=5.1,
    tick label style={font=\scriptsize},
    legend style={font=\scriptsize, at={(0.5,-0.22)}, anchor=north, legend columns=2},
    grid=major, grid style={gray!20},
]
\addplot[blue!70!black, thick, mark=*, mark size=1pt] coordinates
 {(0,5) (50,4.32) (100,4.21) (150,4.1) (200,4.27) (250,4.11) (300,4.56) (350,4.5) (400,4.5) (450,4.38) (550,4.29) (600,4.52) (650,4.18) (700,4.22) (750,4.21) (800,4.26) (850,4.2) (950,4.33) (1000,4.29) (1050,4.4) (1100,4.18)};
\addplot[cyan!60!black, thick, mark=square*, mark size=1pt] coordinates
 {(0,4.83) (50,4.26) (100,4.29) (150,4.21) (200,4.42) (250,3.97) (300,4.67) (350,4.6) (400,4.55) (450,4.54) (550,4.61) (600,4.39) (650,4.47) (700,4.22) (750,4.08) (800,4.36) (850,4.2) (950,4.22) (1000,4.35) (1050,4.42) (1100,4.14)};
\addplot[red!70!black, thick, dashed, mark=triangle*, mark size=1pt] coordinates
 {(0,4.5) (50,4.34) (100,4.29) (150,4.12) (200,4.26) (250,3.98) (300,4.5) (350,4.65) (400,4.55) (450,4.58) (550,4.39) (600,4.29) (650,4.29) (700,4.56) (750,4.29) (800,4.21) (850,4.27) (950,4.45) (1000,4.49) (1050,4.47) (1100,3.8)};
\addplot[orange!80!black, thick, dashed, mark=diamond*, mark size=1pt] coordinates
 {(0,4.5) (50,4.45) (100,4.34) (150,4.23) (200,4.24) (250,3.86) (300,4.33) (350,4.35) (400,4.45) (450,4.42) (550,4.39) (600,4.42) (650,4.47) (700,4.39) (750,4.26) (800,4.26) (850,4.31) (950,4.33) (1000,4.29) (1050,4.49) (1100,3.87)};
\legend{ICE generalist, ICE MoE, Vector generalist, Vector MoE}
\end{axis}
\end{tikzpicture}
\caption{Longitudinal mean score by turn-index bin (1{,}057 probe--checkpoint observations, Dataset C excluded; real per-bin data). The four conditions track closely mid-conversation, but at the final turn-1{,}100 bin---the deep long-horizon regime of Dataset B---both vector conditions drop sharply while both ICE conditions hold.}
\label{fig:longitudinal}
\end{figure}
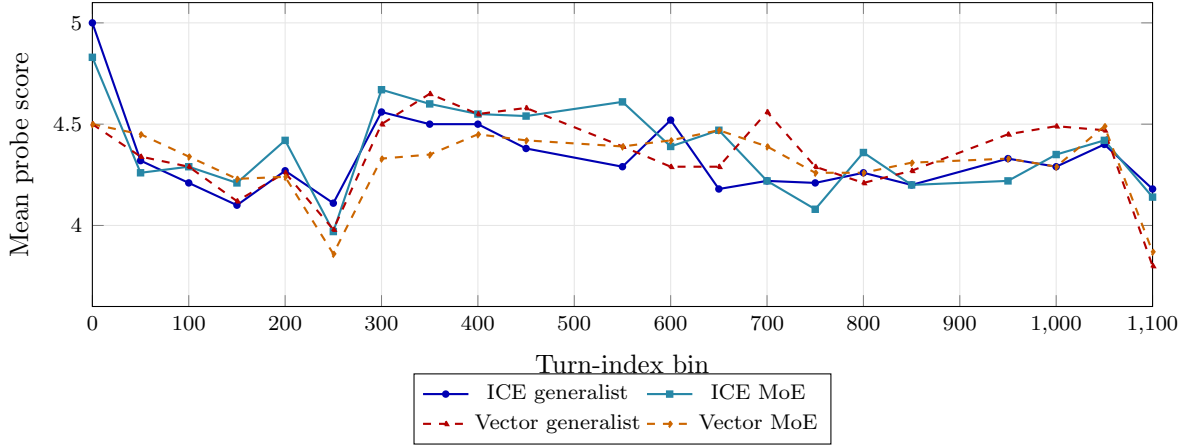

\textbf{Longitudinal behaviour.} \cref{fig:longitudinal} plots mean probe score against turn-index bin. All four conditions track closely through early and middle bins---the tie is visible as four overlapping curves. The separation appears at the deepest bin (turn 1{,}100 of Dataset B): both vector conditions drop sharply, to 3.80 and 3.87, while both ICE conditions hold at 4.18 and 4.14. This final-bin difference is descriptive; it does not identify retrieval reach as the cause, and the bin represents one conversation.

\textbf{Context cost changes with position.} The 6.6\% figure above is an average over conditions that differ sharply, and two mechanisms push against each other. Per-turn \emph{density} favours ICE: the baseline retrieves a fixed top-30, so its injected tokens scale with how large each turn is, without bound. Conversation \emph{length} favours the baseline: ICE's growth cap deliberately widens the budget as turns accumulate ($2{,}000 + 150n$ below 30 turns, then $5{,}000 + 100(n-30)$, then $10{,}000 + 30(n-100)$), while top-30 is indifferent to how long a conversation has run. \cref{tab:token-crossover} contrasts the first and last quartile of each conversation, paired per probe and bootstrapped. These are observational position contrasts: query mix, evidence density, and state also vary with checkpoint.

Dataset D changes from fewer ICE v2 prompt tokens in the first quartile to more in the last. Dataset B's positive cost difference shrinks, so the table does not support a uniform increase in relative cost across datasets. Density dominates Dataset C. The accompanying score changes do not identify the budget's marginal return: query difficulty, state, and context all vary together. A quality--token frontier would require independently varied budgets.

\begin{table}[htbp]
\centering
\small
\caption{Token cost by position within each conversation: first quartile of probes versus last. $\Delta$ is the paired per-probe difference (ICE minus baseline; negative means ICE is cheaper) with a 95\% bootstrap interval, and ``cheaper'' is the share of probes on which ICE injected fewer tokens. Historical record-level intervals; query mix and state vary with position. These are not budget interventions.}
\label{tab:token-crossover}
\begin{tabular}{@{}llrrr@{}}
\toprule
\textbf{Dataset} & \textbf{Position} & \textbf{Paired $\Delta$} & \textbf{95\% CI} & \textbf{ICE cheaper} \\
\midrule
\multirow{2}{*}{A (Creative, 290 turns)} & first quartile & $-8{,}034$ & $[-8{,}650, -7{,}451]$ & 100\% \\
                                          & last quartile  & $-2{,}284$ & $[-2{,}766, -1{,}800]$ & 91\% \\
\midrule
\multirow{2}{*}{B (Long-form, 1{,}119 turns)} & first quartile & $+3{,}766$ & $[+3{,}133, +4{,}394]$ & 18\% \\
                                               & last quartile  & $+1{,}288$ & $[+476, +2{,}075]$ & 35\% \\
\midrule
\multirow{2}{*}{C (Technical, 325 turns)} & first quartile & $-70{,}731$ & $[-79{,}067, -63{,}734]$ & 100\% \\
                                           & last quartile  & $-68{,}854$ & $[-88{,}105, -53{,}696]$ & 100\% \\
\midrule
\multirow{2}{*}{D (Academic, 251 turns)} & first quartile & $-2{,}309$ & $[-3{,}112, -1{,}521]$ & 69\% \\
                                          & last quartile  & $+4{,}994$ & $[+4{,}362, +5{,}609]$ & 5\% \\
\bottomrule
\end{tabular}
\end{table}

\textbf{Routing.} The ordinary-density MoE-versus-generalist score delta is $+0.03$ for ICE and $-0.02$ for vector; the effect is too small to claim in either direction. MoE gives a 3.1\% hallucination reduction under vector and a 0.3\% increase under ICE, and no token savings. The routing condition has a small observed mean difference; we return to its scope in Section~\ref{sec:exercised}.

\textbf{Memory maturity and reliability.} Simulated decay days at the final checkpoint were 93 for Dataset B, 24 for A, 27 for C, 20 for D. Gating failures---probes where the classifier blocked retrieval that was needed---fell from 22 in the pilot to 2 here, the single largest reliability improvement in the mature deployment.

\textbf{Hand-scored subset.} The 72 manual records favour ICE v2 by $+1.21$ score points (historical record-bootstrap CI $[+0.81,+1.61]$). They are included in the reported aggregate and come from one author/scorer. Selection and scorer effects prevent treating this subset as proof that the automated judge underestimates ICE's population advantage.

\begin{table}[htbp]
\centering
\small
\caption{Human-verified slice (72 hand-scored probes, scored by the corpus author). Paired ICE-versus-vector is $+1.21$ (95\% CI $[+0.81, +1.61]$), alongside the near-zero aggregate difference over 1{,}057 probe--checkpoint observations. These manual records also enter the reported aggregate. Single scorer; blind per-probe tournament ranking.}
\label{tab:manual}
\begin{tabular}{@{}lccc@{}}
\toprule
\textbf{Condition} & \textbf{Score} & \textbf{95\% CI} & \textbf{Win \%} \\
\midrule
Vector RAG (generalist) & 3.10 & $[2.75, 3.44]$ & 5.6 \\
Vector RAG (MoE)        & 3.22 & $[2.88, 3.54]$ & 8.3 \\
Full ICE (generalist)   & 4.31 & $[4.11, 4.49]$ & 16.7 \\
Full ICE (MoE)          & 4.24 & $[4.01, 4.43]$ & \textbf{69.4} \\
\bottomrule
\end{tabular}
\end{table}

\subsubsection{Density Stress and the All-Dataset View}
\label{sec:stress}

Dataset C contains architecture documents that frequently exceed 8{,}000 tokens in a single turn. The vector baseline, which always retrieves the top-30 fragments with no per-query budget, injects a mean of 88{,}061 tokens---failed cases reach 100{,}505---overwhelming the model's context window. Of 154 probe--checkpoint observations, 145 receive score 1 (94.2\%); score 1 is a poor-answer/failure category, not a direct telemetry count of context-overflow exceptions. The retrieval subsystem did find relevant information; the generation model could not process the assembled context and produced empty or degenerate responses. ICE, with its dynamic budget capping retrieval, holds a mean of 19{,}953 tokens and a mean score of 4.33.

\begin{table}[htbp]
\centering
\small
\caption{ICE v2 density stress test on Dataset C (154 probe--checkpoint observations). The vector baseline collapses because it has no token-budget cap. ICE survives with a mean score of 4.33 and 77\% fewer tokens injected. The baseline's 0\% hallucination rate is an artefact: 94.2\% of its responses are complete failures, so there is no answer to evaluate.}
\label{tab:stress}
\resizebox{\textwidth}{!}{%
\begin{tabular}{@{}lrrrrrrrr@{}}
\toprule
\textbf{Condition} & \textbf{Score} & \textbf{Tok.} & \textbf{Frags} & \textbf{SPF} & \textbf{TUR} & \textbf{Hall \%} & \textbf{Win \%} & \textbf{Score 1 \%} \\
\midrule
Vector RAG generalist & 1.23 & 88{,}061 & 30 & 0.04 & 0.01 & 0.0* & 1.9 & \textbf{94.2} \\
Vector RAG MoE        & 1.10 & 88{,}061 & 30 & 0.04 & 0.01 & 0.0* & 4.5 & 97.4 \\
Full ICE generalist   & \textbf{4.33} & 19{,}953 & 10 & \textbf{0.43} & \textbf{0.22} & 41.4 & 43.5 & 3.9 \\
Full ICE MoE          & 4.03 & 19{,}953 & 10 & 0.40 & 0.20 & 40.8 & \textbf{50.0} & 16.2 \\
\bottomrule
\end{tabular}%
}
\end{table}

The score distributions are far apart. ICE generalist: 3.9\% at score 1, 5.2\% at 2, 11.7\% at 3, 12.3\% at 4, 66.9\% at 5. The baseline: 94.2\% at score 1, nothing at 2--4, 5.8\% at 5.

ICE v2's 41.4\% hallucination rate remains a substantial limitation. The baseline's near-absence of usable answers makes its zero hallucination rate uninformative about successful-answer faithfulness; that does not excuse ICE errors.

\textbf{Information density as a benchmark dimension.} Turn count alone is an insufficient measure of difficulty. Dataset B is the greatest challenge in \emph{temporal horizon}; Dataset C is a different challenge entirely, \emph{extreme information density}. The benchmark surfaced two distinct stress dimensions, and the second was not designed---it emerged when the first few probes returned errors.

\textbf{What the headline numbers look like with Dataset C included.} \cref{tab:exp2-global} deliberately excludes Dataset C, so the reported tie reflects conditions under which both systems can produce an answer; a 94.2\% failure rate would otherwise dominate any aggregate without saying anything about answer \emph{quality}. So that this choice is not mistaken for cherry-picking, \cref{tab:exp2-alldata} reports the same four conditions over \emph{all} 1{,}211 probe--checkpoint observations.

\begin{table}[htbp]
\centering
\small
\caption{ICE v2 over all four conversations, including the density stress case (1,211 observations). The aggregate combines answer quality with the unbudgeted baseline\'s high failure rate on Dataset C.}
\label{tab:exp2-alldata}
\begin{tabular}{@{}lrrrrrr@{}}
\toprule
\textbf{Condition} & \textbf{Score} & \textbf{Tokens} & \textbf{Frags} & \textbf{SPF} & \textbf{Win \%} & \textbf{Hall.\ \%} \\
\midrule
Vector RAG generalist & 3.87 $\pm$ 1.47 & 29{,}550 & 30.0 & 0.13 & 18.7 & 20.2 \\
Vector RAG MoE        & 3.84 $\pm$ 1.45 & 29{,}550 & 30.0 & 0.13 & 17.3 & 15.7 \\
Full ICE generalist   & \textbf{4.27} $\pm$ 1.09 & 22{,}099 & 19.1 & \textbf{0.22} & \textbf{32.3} & 22.3 \\
Full ICE MoE          & 4.25 $\pm$ 1.03 & 22{,}099 & 19.1 & 0.22 & 31.7 & 22.5 \\
\bottomrule
\end{tabular}
\end{table}

The historical paired record-bootstrap differences are $+0.40$ (95\% CI $[+0.31,+0.49]$) over all data and $+3.10$ ($[+2.85,+3.33]$) on Dataset C. Probe-clustered intervals remain positive but wider (Appendix~\ref{app:clustered}). Both the all-data view and the ordinary-density view are needed: the first includes deployment failures, while the second shows that those failures account for most of the aggregate quality gap.

\subsection{RQ2: Mechanism Fidelity}
\subsubsection{Cumulative Ablation}
\label{sec:ablation}

To understand which mechanisms matter, we ran a cumulative feature buildup on Dataset B (67 probes, fully mature memory state, single pass), constructing the system incrementally from a bare-vector baseline with each step adding exactly one feature and preserving all previous ones. The answering model here is Qwen3-14B-AWQ rather than Experiment 2's 26B generalist; the independent Gemma-4 12B judge is unchanged. Absolute scores are therefore not comparable across experiments, and relative within-probe deltas are the object of interest. \cref{tab:ablation-main} gives the CI-robust steps; the full fifteen-row table, the recency breakdown, and the waterfall chart are in Appendix~\ref{app:ablation}.

\begin{table}[htbp]
\centering
\small
\caption{ICE v2 ablation steps that clear a paired within-probe bootstrap 95\% CI (10{,}000 resamples, 67 probes scored under every condition), with the endpoints for context. Every other step's interval spans zero; the full table is \cref{tab:ablation-full}. The single-leg vector baseline is a reference point, not a buildup step.}
\label{tab:ablation-main}
\begin{tabular}{@{}lrrl@{}}
\toprule
\textbf{Step} & \textbf{Score} & \textbf{Paired $\Delta$} & \textbf{95\% CI} \\
\midrule
bare\_vector             & 3.27 & ---     & --- \\
$+$BM25                  & 2.52 & $-0.74$ & $\LexicalCI$ \quad excludes 0 \\
$+$RRF                   & 3.36 & $+0.82$ & $\FusionCI$ \quad excludes 0 \\
\multicolumn{4}{@{}l@{}}{\emph{\quad nine further steps, all spanning zero (Appendix~\ref{app:ablation})}} \\
full\_ice                & 3.38 & $-0.03$ & $[-0.29, +0.23]$ \\
\midrule
vector\_baseline (ref.)  & 3.42 & ---     & --- \\
\bottomrule
\end{tabular}
\end{table}

\textbf{Adding an unfused lexical leg is harmful} ($-0.74$, $\LexicalCI$). The observed damage is consistent with additional unfused matches admitting distracting context. This buildup shows that adding a lexical leg without fusion can worsen quality; it does not isolate the content responsible for each error.

\textbf{Rank fusion is corrective} ($+0.82$, $\FusionCI$). Adding RRF on top recovers the damage. Its role is precisely that: recovery. Neither RRF-on-vector nor the fully built configuration is statistically distinguishable from the single-leg baseline on answer quality---both contrasts span zero, consistent with the Experiment 2 tie. RRF corrects the damage from adding the lexical leg in this buildup; no general safety guarantee follows. We state this deliberately because the earlier draft of this work claimed RRF was the single most impactful component; the paired intervals do not support that, and the corrected claim is the smaller one.

The remaining steps must be read with care, and Section~\ref{sec:exercised} explains why: several toggled features were not functioning or not varying at the snapshot, so their near-zero deltas measure nothing about the mechanism. The small observed contrasts---which do not establish equivalence---are cluster restriction and session diversification ($\Delta \leq 0.01$). One inconclusive contrast concerns the budget: the dynamic budget's step is $-0.11$, alongside a fill-to-cap policy that allocates more tokens to longer conversations and selects more fragments on Dataset B (21.2 fragments against 11.3 at the pre-budget step). The paired budget-step estimate is $-0.08$ with CI $[-0.36,+0.18]$; it is inconclusive, and does not establish that an alternative budget policy would improve quality.

\subsubsection{Execution, Reach, and Identifiable Effects}
\label{sec:carried}\label{sec:exercised}
The frozen v2 audit finds episodic fragments accounting for 62.3\% of recorded selections, Codex for 3.3\%, and unattributable source records for 34.5\%. The graph concatenates a traversal into one fragment, so fragment share cannot be equated with token share, fact coverage, or utility. The trained NER model was loaded, but that fact alone establishes neither task-specific recall nor graph correctness.

Procedural retrieval is \textbf{defective}: its untyped embedding bind raises a pgvector operator error and returns an empty list. The document leg shares that defect and additionally had no ingested documents. Batch summaries had no eligible stored content; archival retrieval was not exercised. HyDE was disabled in the mature run and constant across the ablation arms rather than controlled by its nominal flag. These are not neutral quality results.

Cluster restriction and model routing executed with small observed differences; they support ``no detected effect in this setting'', not equivalence. Session diversification did not exercise its central cross-conversation cap in the within-conversation replay. Graph and enumeration-fallback contrasts are inconclusive. Current-truth probes can reward correct updates but do not validate as-of graph querying or historical composition. RQ3 supplies the separate cross-session test.

The fusion contrast supports a local causal attribution within this buildup: an unfused lexical leg hurts, and RRF recovers that loss. The density stress case supports the bounded \emph{full-system} configuration against the unbudgeted baseline; it does not isolate the budget from assembly and retrieval. Decay and reinforcement ran but were not independently switched off. The observed fragment reduction belongs to the complete selection-and-assembly condition. Appendix~\ref{app:fidelity} records the component evidence and its limits.

\subsection{RQ3: External Transfer under LongMemEval}
\label{sec:lme-oracle}
\textbf{ICE v2 loses decisively overall in both phases.} Table~\ref{tab:lme-oracle} places pure vector-RAG beside ICE v2 under the same cloud answerer and judge. The evidence-only deficit is already 22.0 percentage points; distractors therefore cannot be its root cause. LSREP's continuing-conversation observations did not transfer to strong endpoint QA across separately supplied sessions.

\begin{table}[htbp]
\centering\small
\caption{Matched LongMemEval accuracy (percent; correct/obtainable verdicts). Both arms use gpt-5.6-luna and Muse Spark 1.3 Contributor judging; these are not official-judge leaderboard scores.}
\label{tab:lme-oracle}
\begin{tabular}{@{}lrrrr@{}}\toprule
 & \multicolumn{2}{c}{Oracle} & \multicolumn{2}{c}{Full-S}\\
Type & ICE v2 & Vector-RAG & ICE v2 & Vector-RAG\\\midrule
Abstention & 83.3 (25/30) & 60.0 (18/30) & 83.3 (25/30) & 63.3 (19/30)\\
Knowledge update & 58.3 (42/72) & 73.6 (53/72) & 51.4 (37/72) & 69.4 (50/72)\\
Multi-session & 29.8 (36/121) & 86.0 (104/121) & 21.5 (26/121) & 74.4 (90/121)\\
Session assistant & 91.1 (51/56) & 94.6 (53/56) & 75.0 (42/56) & 94.6 (53/56)\\
Session preference & 70.0 (21/30) & 70.0 (21/30) & 43.3 (13/30) & 51.7 (15/29)\\
Session user & 78.1 (50/64) & 95.3 (61/64) & 71.9 (46/64) & 93.8 (60/64)\\
Temporal reasoning & 22.8 (29/127) & 42.5 (54/127) & 20.5 (26/127) & 47.2 (60/127)\\
Overall & 50.8 (254/500) & 72.8 (364/500) & 43.0 (215/500) & 69.5 (347/499)\\
\bottomrule\end{tabular}
\end{table}
Oracle's paired difference is $-22.0$ points, CI $[-26.6,-17.4]$ ($n=500$). Full-S's paired difference is $-26.5$, CI $[-31.3,-21.8]$ ($n=499$). One vector judgement is missing: all-500 bounds are 69.4--69.6\% for vector and exactly 43.0\% for ICE v2. Its assignment cannot change the ordering. The full-S paired ICE rate is $215/499$; the displayed ICE marginal is $215/500$. Comparisons of rounded marginal percentages and paired estimates consequently differ slightly.

\paragraph{Category structure.}
Full-S multi-session accuracy is 21.5\% versus 74.4\%, and temporal accuracy is 20.5\% versus 47.2\%. Their paired differences are $-52.9$ points, CI $[-62.8,-43.0]$ ($n=121$), and $-26.8$, CI $[-35.4,-18.1]$ ($n=127$). These expose severe synthesis and temporal-composition failures. They do not identify whether a particular answer failed during construction, retrieval, or reasoning.

Abstention is a descriptive strength: full-S ICE v2 is correct on 25/30 questions versus 19/30 for vector. The paired table contains 18 both-correct, 7 ICE-only, 1 vector-only, and 4 both-wrong outcomes. The +20.0-point percentile-bootstrap interval is $[+3.3,+36.7]$, but there are only eight discordant pairs; a two-sided exact McNemar test gives $p=0.0703$. Oracle abstention is +23.3 points, CI $[+6.7,+40.0]$, with 8 versus 1 discordant pairs. These small exploratory subsets, unadjusted for multiple comparisons and conditional on Muse labels, support cautious reporting of conservative abstention rather than a general superiority claim. Appendix~\ref{app:lme-paired} gives every paired category count and interval.

\paragraph{Phase transitions.}
ICE v2 changes from correct to wrong on 68 questions and wrong to correct on 29, losing 7.8 points over 500. Vector changes 44 and 28 times respectively, losing 3.2 points on its 499 complete phase pairs (the marginal table drops approximately 3.3 points). On the common 499 four-way-complete questions, ICE loses 7.6 points and vector 3.2: the \emph{difference in degradation} is 4.4 points, CI $[-0.2,+9.2]$. Its point estimate widens the deficit, but uncertainty includes no extra degradation. Oracle is not a mathematical upper bound: both arms improve on some individual questions after other histories are added.

\paragraph{Quality and context cost.}
Table~\ref{tab:lme-cost} distinguishes observed input volume from configuration. In full-S, ICE v2 supplies a median of 5 fragments and 2,222 provider input tokens; vector supplies 30 and 11,718. Vector achieves much higher accuracy with substantially more context. These data establish context economy for ICE and a quality--cost trade-off, not an efficiency winner. They do not determine accuracy at matched budgets or whether either system dominates an accuracy--token frontier.

\begin{table}[htbp]
\centering\small
\caption{Recorded costs, median [25th, 75th percentile], 500 answers per row. Fragments are post-selection and supplied to assembly; provider input tokens cover the complete request. Seconds measure answer generation only.}
\label{tab:lme-cost}
\begin{tabular}{@{}llrrr@{}}\toprule
Phase & Arm & Fragments & Input tokens & Generation (s)\\\midrule
Oracle & ICE v2 & 3.0 [3.0, 4.0] & 2,006 [1,758, 2,118] & 2.8 [2.2, 3.4]\\
Oracle & Vector & 11.5 [6.0, 12.0] & 5,529 [3,370, 6,732] & 2.0 [1.6, 2.8]\\
Full-S & ICE v2 & 5.0 [4.0, 6.0] & 2,222 [2,168, 2,289] & 3.1 [2.5, 3.8]\\
Full-S & Vector & 30.0 [30.0, 30.0] & 11,718 [10,437, 12,923] & 2.6 [2.1, 3.5]\\
\bottomrule\end{tabular}
\end{table}
Both phases have 500 non-empty answers per arm and no final answer failure; this does not measure transient retry frequency. Judgement missingness is 0/500 per arm in oracle and 0/500 ICE versus 1/500 vector in full-S. ICE's configured retrieval ceiling is 2,000 estimated tokens; vector's top-30 limit is active and has \emph{no} token ceiling. A retrieval-budget field recorded in vector artifacts belongs to an unused orchestrator object and must not be presented as a vector cap.

The recorded word-based estimate includes system instructions and the question, not only retrieved text. Provider input-token metadata is complete for all 2,000 answers. Per-leg candidate counts, retrieval-only token counts, retrieval latency, and arm-specific memory-construction cost were not persisted at the required granularity. They cannot be recovered exactly from summary fields; the paper reports them as unavailable instead of subtracting an assumed overhead. Selected fragment counts equal the fragments joined by the adapter's assembler, but are not pre-selection candidate counts.

Outcome stratification does not support a simple ``more context fixes the error'' reading. Full-S ICE v2 correct and incorrect answers have nearly identical median inputs (2,224 versus 2,219 tokens); vector's are 11,650 versus 11,855. For full-S multi-session questions, vector answers 90/121 correctly while supplying 30 fragments per answer, yet still fails on 31; ICE answers 26/121. Cost by category and correctness, including dispersion, is preserved in Appendix~\ref{app:lme-cost-strata} and the aggregate artifact. No causal claim follows from these observational strata.

\paragraph{Published-system context.}
Table~\ref{tab:published-context} is external context only. Its rows use different models, prompts, and budgets; no row is ranked against ICE v2. A head-to-head claim would require rerunning the other systems under this stack or rejudging ICE/vector outputs with the official judge.

\begin{table}[htbp]
\centering\small
\caption{Published LongMemEval-S configurations: contextual, not matched to this study. NR means not reported in the cited version; no ICE ranking is implied.}
\label{tab:published-context}
\begin{tabular}{@{}L{2.1cm}L{2.0cm}L{1.7cm}rL{5.1cm}@{}}\toprule
System & Answerer & Judge & Score & Prompt / retrieval / protocol\\\midrule
Zep & GPT-4o-mini & GPT-4o & 63.8\% & Task-specific official judging; Zep retrieval; mean context 1.6K tokens (not a declared cap); supplied S histories~\citep{rasmussen2025zep}\\
Zep & GPT-4o & GPT-4o & 71.2\% & Same reported protocol; system-specific answer prompt; mean context 1.6K tokens~\citep{rasmussen2025zep}\\
Hindsight & GPT-OSS-20B & GPT-OSS-120B & 83.6\% & Retain/recall/reflect; paper-specific judge templates; retrieval budget NR in v1; S, 500 questions~\citep{latimer2025hindsight}\\
Hindsight & Gemini-3 Pro & GPT-OSS-120B & 91.4\% & OSS-120B memory construction; Gemini answer generation; same v1 budget disclosure gap~\citep{latimer2025hindsight}\\
\bottomrule\end{tabular}
\end{table}
\section{Limitations and Implications}
\label{sec:limitations}\label{sec:discussion}
\subsection{Private Data, Repeated Probes, and Reference Construction}
Four datasets provide domain and density variation, but all come from one author. The 1,211 observations are repetitions of 219 questions, not independent examples from 1,211 users or tasks. Probe-clustered sensitivity analysis widens uncertainty, while still conditioning on the same four conversations. Neither analysis supports population-level generalisation. The manually scored subset shares the corpus author, and references were refined through a model-assisted process that may omit relevant details or preserve authoring errors.

LSREP's implementation and specification can be reused with other histories; its exact private-data scores cannot be independently reproduced without those data. The synthetic example demonstrates reference semantics, not empirical validity. A public corpus with auditable revision ledgers and independent annotators would test portability more directly.

\subsection{Replay and Mechanism Boundaries}
Replay cannot model how a generated answer changes a user's next turn, because the future transcript is fixed. Evaluation accesses reinforce ICE v2's memory, so the probe schedule itself affects subsequent state. The historical four-condition harness shares stored memory rather than maintaining independent long-lived states for every arm; results concern that declared shared-state comparison. A separate-arm replay would be a different experiment. Simulated horizons are ageing schedules, not elapsed deployment durations.

Graph output presence does not validate graph truth or historical reasoning. Procedural retrieval was defective, document retrieval unexercised and defective, and several lifecycle paths produced no measurable result. The quality effect of decay, prompt scaffolding, and individual curation steps is not independently identified. The density case compares the full system with an unbudgeted top-30 baseline, so it does not show an advantage over a properly budgeted vector baseline. Session placement, model stack, corpus, and grading differ between LSREP and LongMemEval; their scores cannot be subtracted to estimate a causal session effect.

\subsection{Judging and Cost Boundaries}
The v2 LSREP aggregation includes manual scores and imputation. Its hallucination audit and the uncorrected ablation labels have different status; neither absolute rates nor relative arm differences are guaranteed free of judge bias. The matched LongMemEval study uses a non-official Muse judge. Muse passed a three-case discrimination check; its prior 15-item human-labelled calibration had 73\% agreement in a different setting. Neither is a broad benchmark-specific human-agreement study, and bootstrap intervals do not include judge misclassification or run-to-run writer and answerer variation. The answerer was selected using a small sample from the same public benchmark, which further limits confirmatory interpretation.

Lower token use with lower accuracy is not superior efficiency. A matched-budget sweep, cost at a declared quality threshold, or an accuracy--token frontier is needed to establish a stronger claim. Generation latency excludes retrieval, ingestion, and retries; provider token counts are not dollar costs. The selected cloud stack is also distinct from ICE v2's local-first deployment configuration.

\subsection{Implications for the Next Evaluation}
\label{sec:future-work}
The immediate methodological requirement is to evaluate both evolving state and public endpoint performance, accompanied by an execution audit.
An observed benchmark score belongs to the whole evaluation stack: memory system, answerer, judge, prompt, context budget, and dataset. Future work should rerun fixed memory systems with newer answerers and cross answerer and judge families to test ranking stability. This study does not establish family-related judge bias. The pure vector baseline's 69.5\% full-S score numerically overlaps some contextual published scores, but their different stacks preclude a system ranking. The relevant test of architectural complexity is its incremental benefit over a strong simple baseline under matched conditions, including a budget sweep: overall quality, specific capabilities, and cost at a declared quality level. These are requirements for future comparisons, not results of the present experiments.
 Future experiments should vary prompt scaffolding and budgets independently, use separate state trajectories where conditions mutate memory, and test historical references as well as current truth. Graph quality and cross-session synthesis require explicit evidence-level controls. These are unmeasured research questions. ICE v3 development proceeds independently; no v3 behaviour or repair is used to explain a v2 result here.

\section{Conclusion}
\label{sec:conclusion}
LSREP specifies how to evaluate reconstructed conversational memory over time: declare state transitions and access effects, repeat probes against evolving references, and audit whether the mechanisms named in a result actually ran. ICE v2 provides a substantial architectural case study with typed stores, local operation, retrieval fusion, and bounded prompt construction, together with clear limits on which mechanisms were effective.

The v2 results are regime-dependent. Continuing-use replay shows similar ordinary-density mean scores with fewer fragments, and robustness relative to an unbudgeted baseline under extreme density. Matched LongMemEval shows decisive overall losses, conservative abstention, and severe multi-session and temporal failures. ICE v2 uses less public-benchmark context while giving less accurate answers. The disagreement demonstrates why longitudinal evaluation, mechanism fidelity, and public end-task testing are jointly necessary for a credible memory-system claim.

\section*{Reproducibility and Ethics}
\label{sec:repro}
The canonical manuscript preserves the full archival account. The evaluated system is pinned to \texttt{v2-paper-eval}; the public diagnostic additionally identifies the session adapter and documented run-enablement changes. Aggregate reports, analysis scripts, configuration, and adapter/harness code form the public artifact package. Raw answers, judgements, databases, logs, downloaded benchmark data, credentials, private planning notes, and personal corpora are excluded. Public LongMemEval inputs must be obtained from their original release. Exact private LSREP results are not independently reproducible from the released package.

The private corpora are the sole author's conversational records; no private transcript excerpts are reproduced here. The worked example is synthetic. ICE v2's local storage and explicit memory controls support user inspection, but persistence also increases the consequences of incorrect or sensitive stored information. Local operation is a deployment property, not a proof of privacy or correctness. The matched public diagnostic uses remote generation and judging on public benchmark material. No new personal-data release is needed for the aggregate analyses reported here.

\bibliographystyle{plainnat}
\bibliography{ICE_references}

\appendix

\FloatBarrier
\section{Component Fidelity Audit}
\label{app:fidelity}

\cref{tab:fidelity} is the component-by-component audit summarised in Section~\ref{sec:exercised}, grounded in the code at tag \texttt{v2-paper-eval} and in the frozen Experiment 2 result JSON. It is reproduced in full because the three-way classification---defective, never exercised, contributing---is only checkable if the per-component evidence is visible.

\begin{table}[htbp]
\centering
\footnotesize
\renewcommand{\arraystretch}{1.04}
\caption{Component state at the evaluation snapshot. ``Contributing'' means the component executed and affected retrieved output; ``never exercised'' means the benchmark supplied no input that would engage it; ``defective'' means it executed and failed.}
\label{tab:fidelity}
\begin{tabular}{@{}L{2.6cm}L{2.0cm}L{5.1cm}L{2.9cm}@{}}
\toprule
\textbf{Component} & \textbf{State} & \textbf{Evidence} & \textbf{Claimable as} \\
\midrule
Vector leg (decay-weighted) & Contributing & Typed vector bind present; dominates leg contributions & Validated \\
BM25 leg & Contributing & tsvector leg, no embedding bind needed; with vector, 62.3\% of fragments & Validated \\
RRF fusion & Contributing & $+0.82$ $\FusionCI$ over a lexical and a dense leg & Validated, scoped to two legs \\
Dynamic token budget & Contributing & Stress: unbudgeted baseline 94.2\% score-1; full ICE mean 4.33 & Full-system contrast; budget not isolated \\
Post-fusion curation & Contributing & 32\% fewer fragments; near-zero mean-score difference & Observed aggregate reduction \\
Decay + reinforcement & Contributing & Simulated decay days per conversation (93/24/27/20); access-count and score restoration on every hit & Running; size not isolated \\
Cluster scoping & Contributing & Correct bind; ablation $\approx +0.01$ & Small observed effect \\
Session diversify, dedup, bonuses & Contributing & Post-fusion transforms; keyword boost $+0.12$ & Contributing \\
Codex graph & Contributing, under-weighted & 3.3\% of fragments, one concatenated traversal fragment; trained NER checkpoint loaded & Observed output; utility unconfirmed \\
Classifier + gate & Contributing & Gating failures 22 $\to$ 2 & Validated, with the override caveat \\
MoE routing & Small observed effect & Ran on every probe; $\Delta +0.03$ / $-0.02$ (ordinary density) & No detected effect \\
\midrule
Procedural leg & \textbf{Defective} & Untyped array bind $\Rightarrow$ \texttt{vector <=> double precision[]} $\Rightarrow$ rollback $\Rightarrow$ \texttt{[]}; 0.0 fragments & Bug; no claim \\
\midrule
Document (RAG) leg & Never exercised & No documents ingested; store empty; leg also shared the bind defect & No claim either way \\
Batch summaries & Never exercised & Bind correct; nothing decayed far enough to batch; store empty & No claim either way \\
Cold storage & Never exercised & Replay horizon never reached the archival regime & No claim either way \\
Temporal / timeline retrieval & Never exercised & All 1{,}211 probe--checkpoint observations ask for current truth & No claim either way \\
Cross-conversation retrieval & Not exercised by LSREP & Every LSREP probe scoped to one conversation; LongMemEval oracle later exposes a fresh-session deficit & Within-conversation claim only \\
HyDE rewrite & Never isolated & Exp 2: disabled. Exp 3: gated by context-reliance, so active in \emph{all} arms including bare vector; the flag toggled nothing & No claim either way \\
\bottomrule
\end{tabular}
\end{table}

The 34.5\% unattributable source records are an instrumentation limitation. Graph fragment granularity and a loaded NER checkpoint do not establish graph correctness or content share.

\FloatBarrier
\section{Historical ICE v1 Pilot}
\label{app:exp1}

\begin{table}[htbp]
\centering
\small
\caption{How the two LSREP instantiations differ. The shared skeleton---state reconstruction, synchronous worker trigger, output-level judging by the same independent judge---is described in Section~\ref{sec:lsrep}.}
\label{tab:exp-diff}
\begin{tabular}{@{}L{2.8cm}L{4.6cm}L{4.6cm}@{}}
\toprule
\textbf{Dimension} & \textbf{Experiment 1 (pilot)} & \textbf{Experiment 2 (mature)} \\
\midrule
System maturity & Vector + BM25 legs; MiniLM classifier; static budget; uncapped sliding window; graph non-functional & All legs enabled; Qwen3 classifier + routing override; dynamic budget; cluster-scoped retrieval \\
Corpus & 18 conversations (isolated checkpoints) & 4 long-horizon conversations (full-lifespan replay) \\
Split range & $[\max(10,0.30L),\,0.95L]$, 3 splits/conv & Length-adaptive checkpoints (8--20), even spacing with jitter \\
Probes & Hand-written (8--30/conv) & Auto-generated, leg-aware (147) plus 72 inherited hand-written \\
Ground truth & Single retrieval-assisted pass, \emph{frozen} & Three-stage forensic regeneration + forward propagation, \emph{evolving} \\
Checkpoint state & Independent per checkpoint & Preserved across checkpoints; lifecycle runs between them \\
Probe evaluation & Each probe once & Incremental: re-scored at every later checkpoint \\
Scoring & Static correctness & Temporal-aware + inference credit; hallucination flags audited for false positives \\
Conditions & Six (incl.\ sliding-window controls) & Four (controls dropped) \\
\bottomrule
\end{tabular}
\end{table}

\textbf{System under test.} The pilot evaluated an immature build with only the vector and BM25 legs meaningfully active. The classifier used a frozen MiniLM backbone; retrieval used static per-leg limits, a static 5{,}000-token global budget, and an uncapped 10-turn sliding window; the knowledge graph effectively did not exist---a regex tagger and an under-powered 1.5B extractor produced almost no usable edges---so the graph, procedural, and document legs contributed nothing and the system reduced to essentially the same vector-plus-BM25 retrieval as the baseline; cluster-scoped retrieval did not exist yet.

\textbf{Protocol instantiation.} Conversations shorter than 10 turns were excluded, reducing the corpus from 42 to 18. Split points were drawn from $[\max(10,0.30L),\,0.95L]$, three per conversation with a fixed seed; each split defined a historical block and a 10-turn future reference block. Probes were hand-written, 8--30 per conversation, and probes from earlier splits were re-asked at later ones. Ground truth used a single retrieval-assisted pass, frozen thereafter. Six conditions ran per probe.

\textbf{The token-accounting audit.} After the pilot completed, an audit found that the reported context size for full ICE included retrieved fragments but \emph{omitted} the recent-turn sliding window incorporated during prompt assembly. Generated answers, scores, rankings, and hallucination labels were unaffected---the window was correctly supplied to the model at inference time; only the reported efficiency statistics were wrong. We replayed the sliding-window assembly for every checkpoint and added the omitted tokens. The correction did not change the qualitative findings, but it invalidated any claim of superior token efficiency: corrected, full ICE consumes \emph{more} total tokens than the baseline at roughly equal quality. All pilot numbers reported here are corrected values.

\begin{table}[htbp]
\centering
\small
\caption{Experiment 1 (immature system), 657 probes, corrected token counts. After correcting the sliding-window under-count, ICE used more tokens than the vector baseline, scored slightly lower, hallucinated more, and had a worse TUR. These historical ICE v1 results do not measure frozen ICE v2.}
\label{tab:exp1}
\begin{tabular}{@{}lrrrrr@{}}
\toprule
\textbf{Condition} & \textbf{Score} & \textbf{Tokens} & \textbf{Hall.\ (\%)} & \textbf{TUR} & \textbf{Win \%} \\
\midrule
Control baseline (generalist) & 3.05 $\pm$ 1.61 & 3{,}998  & 74.1 & 0.76 & 11.3 \\
Control MoE                   & 3.01 $\pm$ 1.61 & 3{,}998  & 74.4 & 0.75 & 7.6  \\
Vector RAG (generalist)       & \textbf{4.06} $\pm$ 1.31 & \textbf{5{,}847}  & \textbf{64.7} & \textbf{0.69} & 22.9 \\
Vector RAG (MoE)              & 4.05 $\pm$ 1.32 & 5{,}847  & 66.7 & 0.69 & 15.6 \\
Full ICE (generalist)         & 4.04 $\pm$ 1.33 & 8{,}586  & 69.6 & 0.47 & 22.0 \\
Full ICE (MoE)                & 3.96 $\pm$ 1.33 & 8{,}586  & 72.3 & 0.46 & 20.6 \\
\bottomrule
\end{tabular}
\end{table}

The headline numbers were a defeat: 4.04 against 4.06, 47\% more tokens, more hallucination, a worse TUR. The picture brightened only on the longitudinal axis---193 tracked question curves showed scores improving over time on repeated probes, demonstrating memory accumulation. Several subsystems were broken or immature: only three decay cycles had run; the graph effectively did not exist; 22 gating failures meant the classifier was actively blocking retrieval on probes that needed it; and the budget was static.

\textbf{Lessons.} A short replay with incomplete consolidation tests that particular state, not mature lifecycle behaviour. Honest token counting is essential---the under-count would have made ICE look more efficient than it was. And a broken subsystem can drag down an entire system, which is itself a finding about the fragility of multi-component architectures and a direct ancestor of the fidelity audit in Appendix~\ref{app:fidelity}.

\FloatBarrier
\section{ICE v2 Full Ablation Results}
\label{app:ablation}

\begin{table}[htbp]
\centering
\small
\caption{Cumulative feature addition on Dataset B (67 probes), full buildup. Paired within-probe bootstrap, 10{,}000 resamples. Only $+$BM25 and $+$RRF clear a 95\% CI; every other step spans zero. Steps whose feature was defective, never exercised, or never isolated at the snapshot ($+$procedural, $+$batch summaries, $+$HyDE) are \emph{not} interpretable as neutral findings, and $+$Codex is under-weighted by construction (Section~\ref{sec:exercised}). The vector baseline is a reference point, not a buildup step.}
\label{tab:ablation-full}
\resizebox{\textwidth}{!}{%
\begin{tabular}{@{}lrrrlrrr@{}}
\toprule
\textbf{Step} & \textbf{Score} & \textbf{Step $\Delta$} & \textbf{Paired $\Delta$} & \textbf{95\% CI} & \textbf{SPF} & \textbf{Tokens} & \textbf{Frags} \\
\midrule
bare\_vector             & 3.27 & ---      & ---     & ---                & 0.23 & 14{,}873 & 14.0 \\
$+$BM25                  & 2.52 & $-0.75$  & $-0.74$ & $\LexicalCI$   & 0.17 & 14{,}868 & 15.2 \\
$+$RRF                   & 3.36 & $+0.84$  & $+0.82$ & $\FusionCI$   & 0.29 & 14{,}870 & 11.7 \\
$+$HyDE                  & 3.39 & $+0.03$  & $+0.03$ & $[-0.13, +0.21]$   & 0.29 & 14{,}870 & 11.6 \\
$+$cluster\_restrict     & 3.40 & $+0.01$  & $+0.01$ & $[-0.16, +0.21]$   & 0.30 & 14{,}911 & 11.3 \\
$+$session\_diversify    & 3.40 & $+0.00$  & $+0.00$ & $[-0.22, +0.22]$   & 0.30 & 14{,}911 & 11.2 \\
$+$codex                 & 3.43 & $+0.03$  & $+0.03$ & $[-0.21, +0.25]$   & 0.30 & 14{,}912 & 11.3 \\
$+$MERA                  & 3.22 & $-0.21$  & $-0.21$ & $[-0.43, +0.01]$   & 0.28 & 14{,}912 & 11.4 \\
$+$procedural            & 3.39 & $+0.17$  & $+0.16$ & $[-0.07, +0.40]$   & 0.30 & 14{,}912 & 11.4 \\
$+$batch\_summary        & 3.41 & $+0.02$  & $-0.02$ & $[-0.24, +0.23]$   & 0.30 & 14{,}912 & 11.3 \\
$+$dynamic\_budget       & 3.30 & $-0.11$  & $-0.08$ & $[-0.36, +0.18]$   & 0.16 & 24{,}489 & 21.2 \\
$+$sliding\_window       & 3.32 & $+0.02$  & $-0.02$ & $[-0.24, +0.23]$   & 0.16 & 27{,}763 & 21.3 \\
$+$keyword\_boost        & 3.44 & $+0.12$  & $+0.08$ & $[-0.22, +0.37]$   & 0.23 & 27{,}769 & 15.1 \\
full\_ice                & 3.38 & $-0.06$  & $-0.03$ & $[-0.29, +0.23]$   & 0.22 & 27{,}768 & 15.0 \\
\midrule
vector\_baseline (ref.)  & 3.42 & ---      & ---     & ---                & 0.11 & 16{,}373 & 30.0 \\
\bottomrule
\end{tabular}%
}
\end{table}

Headline contrasts under the same paired bootstrap: the BM25 damage is $-0.74$ $\LexicalCI$ and the RRF rescue $+0.82$ $\FusionCI$, both excluding zero; RRF-versus-bare is $+0.09$ $\FusionBareCI$, full ICE versus bare is $+0.09$ $\FullBareCI$, and full ICE versus the single-leg baseline is $-0.06$ $\FullVectorCI$, all spanning zero.

Two steps warrant individual comment. \textbf{MERA}, the enumeration fallback for category queries ($-0.21$, $[-0.43,+0.01]$, not significant), fires only when entity extraction returns nothing; the audit reports a narrow trigger subset, so this delta rests on a small effective sample and is a downstream indicator of extraction gaps rather than an independent evaluation. \textbf{Codex} ($+0.03$) was live but emitted its entire traversal as one fragment, so fragment counts alone do not measure graph quality.

\begin{figure}[htbp]
\centering
\begin{tikzpicture}
\begin{axis}[
    width=\columnwidth, height=5.8cm,
    ybar,
    bar width=8pt,
    ymin=2.3, ymax=3.7,
    ylabel={\small Mean score after step},
    ylabel near ticks,
    symbolic x coords={bare,+BM25,+RRF,+HyDE,+clust,+sess,+codex,+MERA,+proc,+batch,+budget,+slide,+kw,full ICE},
    xtick=data,
    x tick label style={rotate=55, anchor=east, font=\scriptsize},
    ytick={2.4,2.6,2.8,3.0,3.2,3.4,3.6},
    tick label style={font=\scriptsize},
    enlarge x limits=0.05,
    grid=major, grid style={gray!20},
    every axis plot/.append style={fill=blue!55!black, draw=blue!30!black},
]
\addplot coordinates {
 (bare,3.27) (+BM25,2.52) (+RRF,3.36) (+HyDE,3.39) (+clust,3.40)
 (+sess,3.40) (+codex,3.43) (+MERA,3.22) (+proc,3.39) (+batch,3.41)
 (+budget,3.30) (+slide,3.32) (+kw,3.44) (full ICE,3.38) };
\draw[dashed, red!70!black, thick] (axis cs:bare,3.42) -- (axis cs:full ICE,3.42)
  node[pos=0.12, above, font=\scriptsize, red!70!black] {vector baseline 3.42};
\end{axis}
\end{tikzpicture}
\caption{Feature ablation buildup (Dataset B, 67 probes). Each bar is the cumulative mean score after adding one feature to all previous ones. The only two movements clearing a paired bootstrap CI are the $+$BM25 drop and the $+$RRF recovery. The dashed line marks the single-leg vector baseline, which is not a buildup step; the full-build contrast has uncertainty spanning zero.}
\label{fig:ablation-waterfall}
\end{figure}
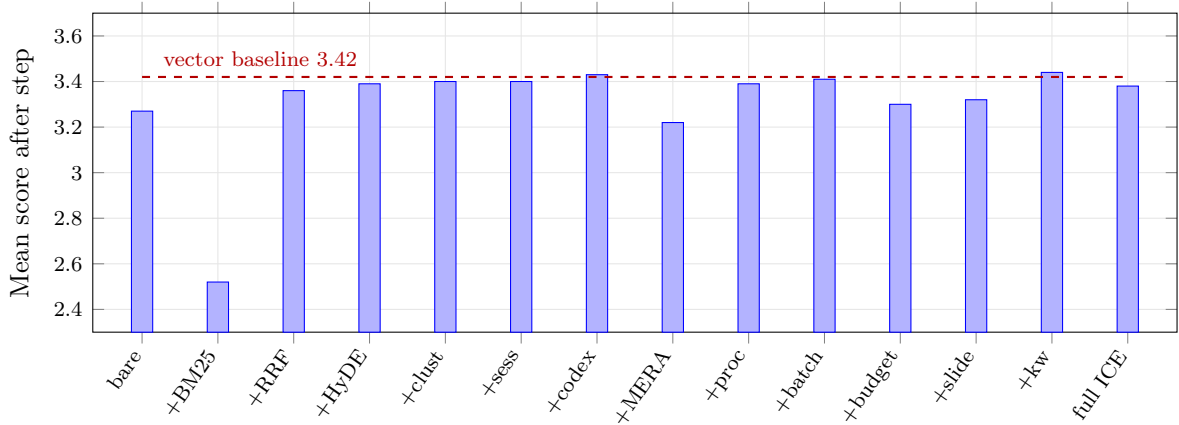

\begin{table}[htbp]
\centering
\small
\caption{Recency effect by origin split (fact age). Probes grouped by the turn index at which they were created. Recency $\Delta$ is mean(late) $-$ mean(early). The fused configurations show positive recency---newer-origin probes have higher scores at the fixed final snapshot---while the single-leg baseline is slightly negative.}
\label{tab:ablation-recency}
\begin{tabular}{@{}lrrrr@{}}
\toprule
\textbf{Condition} & \textbf{Early (0--400)} & \textbf{Mid (400--800)} & \textbf{Late (800--1200)} & \textbf{Recency $\Delta$} \\
\midrule
bare\_vector       & 3.00 & 2.83 & 3.42 & $+0.42$ \\
$+$BM25            & 2.20 & 3.05 & 2.49 & $+0.29$ \\
$+$RRF             & 3.00 & 2.83 & 3.54 & $+0.54$ \\
$+$keyword\_boost  & 2.90 & 3.28 & 3.60 & $+0.70$ \\
full\_ice          & 3.10 & 3.25 & 3.47 & $+0.37$ \\
vector\_baseline   & 3.50 & 3.15 & 3.47 & $-0.03$ \\
\bottomrule
\end{tabular}
\end{table}

\FloatBarrier
\section{ICE v2 Detailed LSREP Breakdowns}

\begin{table}[htbp]
\centering
\small
\caption{Evaluation metrics. Each is computed per probe and aggregated across the benchmark.}
\label{tab:metrics}
\begin{tabular}{@{}L{2.2cm}L{6.2cm}L{4.2cm}@{}}
\toprule
\textbf{Metric} & \textbf{Definition} & \textbf{What it measures} \\
\midrule
Score (1--5) & Mean absolute score from the judge & Answer quality \\
SPF & $\mathrm{mean\ score} / \mathrm{mean\ fragments}$ & Score/fragment ratio \\
TUR & $\mathrm{mean\ score} / (\mathrm{mean\ tokens}/1000)$ & Score/token ratio \\
Win rate (\%) & Blind tournament: all four conditions' answers for a probe are anonymised, shuffled, and ranked best-to-worst in a single pass, independent of absolute scoring; the win rate is the share of appearances placed first & Preference under shuffled presentation \\
Hallucination (\%) & Share of probes where the judge detects information not present in the conversation & Faithfulness \\
Fragment count & Mean fragments injected per probe & Retrieval volume \\
Recency $\Delta$ & $\mathrm{mean\ score}(\mathrm{late\ bins}) - \mathrm{mean\ score}(\mathrm{early\ bins})$ & Descriptive temporal contrast \\
\bottomrule
\end{tabular}
\end{table}
\label{app:exp2-detail}

The four-dataset main-text table is \cref{tab:exp2-perconv}; additional distributions follow.

\begin{table}[htbp]
\centering
\small
\caption{Score distribution (1{,}057 probe--checkpoint observations, Dataset C excluded). The mean-score tie masks a structural difference: ICE-MoE has the lowest score-1 rate but shifts mass from 5 to 3--4, producing lower-variance output.}
\label{tab:exp2-score-dist}
\begin{tabular}{@{}lrrrr@{}}
\toprule
\textbf{Score} & \textbf{ICE gen (\%)} & \textbf{ICE MoE (\%)} & \textbf{Vec.\ gen (\%)} & \textbf{Vec.\ MoE (\%)} \\
\midrule
1 (poor)       & 3.0 & 1.5 & 3.8 & 2.4 \\
2              & 5.9 & 4.3 & 4.3 & 3.9 \\
3 (acceptable) & 13.3 & 16.6 & 14.8 & 19.5 \\
4 (good)       & 18.1 & 19.7 & 17.2 & 16.3 \\
5 (excellent)  & 59.7 & 58.0 & 60.0 & 58.0 \\
\bottomrule
\end{tabular}
\end{table}

\begin{table}[htbp]
\centering
\small
\caption{Temporal score quality (1{,}057 probe--checkpoint observations, Dataset C excluded). Buckets: Good (4--5), OK (3), Poor (1--2).}
\label{tab:exp2-temporal}
\begin{tabular}{@{}lrrr@{}}
\toprule
\textbf{Condition} & \textbf{Good (4--5) \%} & \textbf{OK (3) \%} & \textbf{Poor (1--2) \%} \\
\midrule
Vector RAG generalist  & 77.2 & 14.8 & 8.0 \\
Vector RAG MoE         & 74.3 & 19.5 & 6.2 \\
Full ICE generalist    & \textbf{77.8} & 13.3 & 8.9 \\
Full ICE MoE           & 77.7 & 16.6 & 5.8 \\
\bottomrule
\end{tabular}
\end{table}

\FloatBarrier
\section{Frozen ICE v2 Implementation Reference}
\label{app:implementation}

This appendix collects operational detail supporting Section~\ref{sec:architecture} but not required to follow the argument. A complete technical reference---the full training pipeline for the classifier and NER models, the controlled relation vocabulary in its entirety, GPU resource management, and the idempotency architecture---is the archived technical report in the repository, which describes the system at the tagged evaluation snapshot.

\subsection{Classification Engine}
\label{app:classifier}

The engine produces, per user turn, a set of topic tags, a set of intent tags, and one context-reliance label. These drive every downstream decision: which legs are weighted, how the budget is split, whether the wide-net fallback fires, and which model the router selects.

The learned head is deliberately tiny---a $384 \rightarrow 128 \rightarrow 25$ multi-layer perceptron with ReLU and 0.3 dropout. Its input is a single 384-dimensional embedding from a frozen \texttt{Qwen3-Embedding-0.6B} encoder; the shared linear head is sliced at inference into three blocks: 11 topic labels, 11 intent labels, and 3 context-reliance classes. Topic and intent decode multi-label (sigmoid, threshold 0.3, argmax fallback); context-reliance decodes single-label.

The rule-based pre-classifier computes five density signals in $[0,1]$ from the raw prompt---code density, sentiment density, meta density (references to the model itself), noise density, and reference density (anaphoric terms)---and evaluates five rules in order, returning the first that fires or none, in which case the learned head runs. The reference rule is special: it returns empty topic and intent tags but forces Long\_Term\_Memory, and the orchestrator then runs the head only to obtain topic and intent. A two-tier threshold on that rule (0.2 for short conversations, 0.1 past ten turns) is the engine's primary long-conversation memory bias.

\textbf{Override rules} run in evaluation order after either path: memory immutability, so once Long\_Term\_Memory is set nothing may downgrade it; a topic rule promoting creative turns to Long\_Term\_Memory; and a rule promoting technical turns containing anaphora. A second, API-level bias lives outside the classifier: past ten turns, or below 0.95 confidence, a Zero\_Shot label is upgraded before retrieval runs. Section~\ref{sec:limitations} discusses what this costs.

\textbf{Context-aware classification.} When a conversation ID is supplied, the learned path queries the last three episodic turns for that conversation (max 500 words), preferring summaries and falling back to the first 150 words of raw text, and prepends that context to the prompt before embedding.

\subsection{Codex Write Path and Retrieval}
\label{app:codex}

The graph spans four tables: entities (canonical name, aliases, properties JSONB, auto-regenerated context payload, embedding); edges (typed relations with strength, a confidence flag in \{pending, active\}, \texttt{valid\_from}, and \texttt{valid\_until}, where NULL means treated as current by the store); an append-only event log; and snapshots.

The triplet write path is the heart of temporal versioning. Entity resolution is two-stage---exact canonical-name match, then alias match, else create with a deterministic UUIDv5. The write branch then depends on the relation bucket: a property observation expires prior active edges of the same (source, relation) pair, writes a new edge, and overwrites the JSONB key; a reinforcement of an existing non-property edge increases strength and promotes from pending to active at strength $\geq 2$; a new relation between the same pair expires the old one only if the old relation is not multi-valued. Every state change emits an event, and the entity's context payload is rebuilt from the property map plus the ten most recent active outgoing edges. The event log is compacted by a worker that snapshots any entity exceeding 100 uncompacted events---textbook event sourcing, with the live edge table as current state, the log as audit trail, and snapshots as bounded replay.

\textbf{Micro-NER} is a BIO tagger (a $384 \rightarrow 128 \rightarrow 64 \rightarrow 3$ MLP over per-token embeddings). When the trained model is unavailable the system falls back to a capitalisation regex minus a stoplist; at the Experiment 2 snapshot the trained model was present and loaded. \textbf{Vector fuzzy matching} embeds the extracted entity strings, scans entity embeddings, and takes the best above a cosine threshold of 0.85---the mechanism that resolves a slightly misspelled proper noun to its canonical entity.

The retrieval leg extracts prompt entities, resolves them by fuzzy matching, optionally restricts to a conversation scope, and performs a BFS traversal to depth 3 over edges where \texttt{valid\_until IS NULL}, appending a labelled context payload per visited entity. \textbf{MERA}, the enumeration fallback for category queries, activates only when NER extracts no entities \emph{and} the prompt contains both a category trigger and an enumeration hint; candidates are deduplicated and ranked by 30-day mention count.

\subsection{Codex Limitations in Detail}
\label{app:codex-limits}

\textbf{Extraction granularity.} The frozen v2 extractor uses 6,000-token chunks with overlap. The audit records extraction and representation limitations, but this study does not establish an optimal chunk size or a general capacity limit for 4B models. Such claims require a controlled extraction study.

\textbf{Inert confidence and strength.} Edges carry a confidence flag and a decaying numeric strength, but both are under-used at retrieval time. Traversal follows any edge with \texttt{valid\_until IS NULL} regardless of either, and the only effect of an active confidence flag is a coarse $1.5\times$ fragment-level score boost. There is no reinforcement loop strengthening edges on retrieval, so frequently referenced facts do not rise in the ranking and traversal treats weak edges as equal to strong ones.

\textbf{Lexical versus semantic entity matching.} Retrieval resolves entities by canonical name or alias plus fuzzy matching over the context-payload embedding rather than the entity name. A user who establishes ``The Obsidian Citadel'' and later asks ``where is the main fortress located?'' gives the resolution step no lexical or embedded anchor for ``fortress,'' though a human reader resolves it immediately. This is a general limitation of entity-centric retrieval in free-form conversation: it assumes users refer to entities by canonical names or close aliases, which natural dialogue does not honour.

\subsection{Prompt Assembly}
\label{app:assembly}

The assembler emits a message list in a deliberately stable-prefix order to maximise cache reuse: (i)~a fixed system message with an inline persistent-context block rendering each active memory slot; (ii)~recent turns as alternating user/assistant pairs under dynamic per-turn word caps; (iii)~a single user message headed with the retrieved context, optionally tagged with cluster names when a cluster scope is populated; (iv)~a single assistant acknowledgement acting as a boundary marker, so the model treats the final user message as the live question rather than another history turn; and (v)~the live question. Because the system message, slots, and most of the recent-turn prefix change slowly, most of the prefix key/value tensors are in principle reusable across consecutive requests; no prompt-cache benefit was measured in this study.

The orchestrator budgets selected fragment text. The frozen HTTP wrapper attempts an additional word check, but computes only the system-message and question words, omitting recent and retrieved messages. Its nominal 4,096-token-derived threshold is therefore not a valid total-prompt ceiling. The experimental adapters call retrieval and assembly directly and do not use this wrapper check. No secondary full-context guarantee is claimed.

\textbf{Memory slots} are prepended verbatim to every prompt. Seven slot names are server-enforced: persona, user preferences, tool guidelines, project context, guidance, pending items, and session patterns. Slots are written through four paths: direct user update, batch initialisation, reflection-proposed but user-gated updates (high-stakes slots land in a review queue), and reflection-applied updates to pending items only.

\textbf{Context clusters} are conversation-scoped topical groupings. The centroid of member turn embeddings, renormalised to unit length after every change, is a 384-dimensional vector; membership is many-to-many. The clustering worker assigns each unassigned turn to the best existing cluster---combining embedding similarity, a bonus per shared entity, and tag overlap---or opens a new cluster when no candidate clears a 0.6 similarity threshold.

\subsection{Background Workers and Infrastructure}
\label{app:workers}

\begin{table}[htbp]
\centering
\small
\caption{Background workers. Configured schedules, not evidence that every worker produced useful output in the evaluation.}
\label{tab:workers}
\begin{tabular}{@{}L{3.2cm}L{2.6cm}L{7.4cm}@{}}
\toprule
\textbf{Worker} & \textbf{Trigger} & \textbf{Function} \\
\midrule
Post-flight evaluator & every turn & lossless detection, summary, dispatch extractors \\
Codex extractor       & lossless turns  & triplet extraction and versioned writes \\
Procedural extractor  & every turn      & pattern detection, reinforcement or insertion \\
Episodic decay        & 1.5 h           & access-weighted decay, archive at 0.1, cold-store at 0.05 \\
Codex decay           & 1.5 h           & edge strength decay, demote below strength 0.3 \\
Procedural decay      & 1.5 h           & deactivation after 180 d with $<$ 3 reinforcements \\
Clustering            & 30 min          & assign turns; merging is separately callable \\
Reflection            & 2 h             & session synthesis, pattern crystallisation, slot proposals \\
Batch summariser      & 2 h             & 50-turn batches, preservation-prompt summaries \\
Sentinel monitor      & 30 min          & declarative rules: threshold, absence, and others \\
Fine-tune             & weekly          & retrain the classifier head on curated labels \\
Compaction            & manual          & snapshot entities with $\geq$ 100 uncompacted events \\
Drop zone             & filesystem      & ingest documents into the document store \\
Codex inject watcher  & filesystem      & ingest structured files into the graph with manual confidence \\
\bottomrule
\end{tabular}
\end{table}

All GPU-touching workers gate on a utilisation threshold polled from the driver and, in shared-model mode, additionally on a user-activity key with a short idle window. Retry countdowns are fixed per class. Extraction tasks use idempotency keys; this is not a verified exactly-once guarantee for every worker.

The system is a single service exposing an OpenAI-compatible chat-completions endpoint plus routers for memory slots, user control, and the model registry, backed by PostgreSQL with pgvector---one 384-dimensional vector column per vector table, because the embedder is shared---and a worker fleet over a message broker. Configuration is a typed settings object read from the environment. Server-sent event types (\texttt{classified}, \texttt{retrieval}, \texttt{context\_ready}, \texttt{generating}, \texttt{degraded}) drive an observability panel and downstream monitoring. Classifier fine-tuning writes a timestamped checkpoint but does not auto-promote it, so production promotion requires a deliberate swap.

\FloatBarrier
\section{Detailed ICE v2 Retrieval Configuration}
\subsection{Hybrid Retrieval Orchestrator}
\label{app:orchestrator}

The orchestrator is the core of pre-flight and the component this study measures most directly.

\textbf{Retrieval legs.} Six legs are defined: \textbf{BM25} over a Postgres tsvector of raw and summary text, filtered on decay score and archival status; \textbf{vector} with decay weighting, scoring $(1 - (\text{embedding} \Leftrightarrow \text{query}))\cdot\text{decay\_score}$, which is what distinguishes this leg from pure semantic search---a highly similar but decayed turn is down-ranked; \textbf{Codex graph traversal}; \textbf{procedural} behind a hard intent gate; \textbf{RAG}, triple-gated and global rather than conversation-scoped; and \textbf{batch summaries}, conversation-scoped. Section~\ref{sec:exercised} reports which of these actually contributed during the evaluation, and why.

\textbf{Dynamic leg weighting.} Base weights are $\{$bm25 0.8, vector 1.0, codex 0.5, procedural 0.2, rag 1.0$\}$. Five intent profiles override them---Factual\_Retrieval boosts vector and demotes codex; Generation, Ideation, and Open\_Exploration do the reverse---and two cumulative topic overrides then apply. This is the mechanism by which inferred intent conditions retrieval, and it is the least-explored lever in the system.

\textbf{Reciprocal Rank Fusion} combines the legs:
\begin{equation}
\mathrm{score}_{\mathrm{RRF}}(f) = \sum_{\ell \in \mathrm{legs}} \frac{\alpha_\ell}{k + \mathrm{rank}_\ell(f)}, \quad k = 60,
\end{equation}
where $\alpha_\ell$ is the blended leg weight and $\mathrm{rank}_\ell(f)$ starts at 1. Fragments are deduplicated \emph{during} fusion: the first occurrence is registered and later occurrences add to its score, so agreement across legs is rewarded. Because RRF consumes ranks rather than scores, heterogeneous legs need no score normalisation to be combined---a plausible explanation for the observed corrective effect (Section~\ref{sec:ablation}).

\textbf{Post-fusion curation} applies a sequence of transforms; their combined condition has the reported fragment reduction, without individual causal attribution. Additive bonuses reward keyword overlap and length and penalise very short fragments; a recency bonus favours the most recent decile but is \emph{skipped} for creative turns, because recent meta-discussion is noise for narrative work. Session diversification caps foreign conversations at three fragments each while leaving the active conversation uncapped. Deduplication hashes fragment text. A two-phase token budget then prioritises source-type diversity---the best fragment of each source type is admitted first if it fits---before greedily filling the remainder. Finally the strengthening step writes back the access-count and decay-score restoration described above.

\textbf{Cluster-scoped retrieval} restricts episodic search to the most relevant topical clusters, falling back to global search when no cluster clears a threshold.

\textbf{Dynamic token budget.} A total context budget of 23{,}000 tokens less an overhead reserve leaves 21{,}200 available. A length-based recent-window fraction, reduced by token density and shifted by intent and topic, splits this between recent turns and retrieval, and a \emph{growth cap} bracketed by turn count prevents long conversations from over-allocating. Leftover budget is intentionally left unused: this constrains the retrieved set and, under extreme per-turn density, averts the context overflow that sinks an unbudgeted baseline (Section~\ref{sec:stress}). On ordinary turns ICE's total token count is comparable to the baseline's---the curation shows up as fewer \emph{fragments}, not fewer tokens. A configurable subclass of the orchestrator exposes flags that switch legs and post-processing steps on and off without redefining classification or fusion; Section~\ref{sec:ablation}'s buildup uses that mechanism.

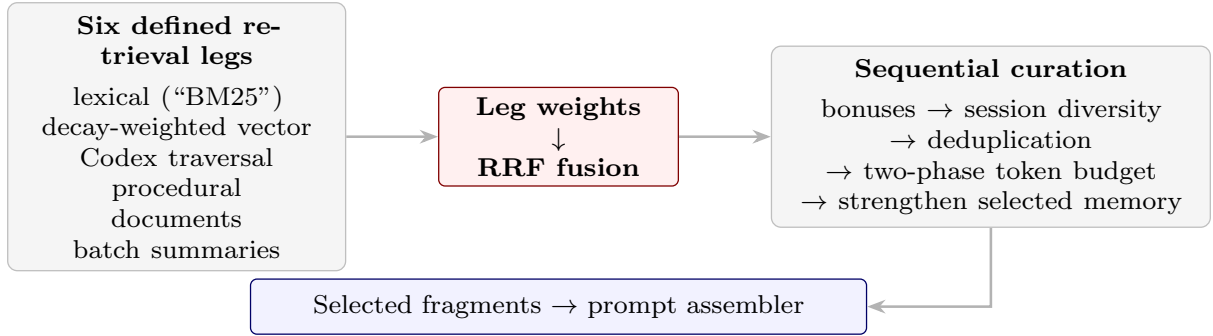
\begin{figure}[t]
\centering
\resizebox{\columnwidth}{!}{%
\begin{tikzpicture}[node distance=4mm]
\node[icephase, text width=34mm] (legs) {Six defined retrieval legs\\[3pt]
\normalfont lexical (``BM25'')\\decay-weighted vector\\Codex traversal\\procedural\\documents\\batch summaries};
\node[icecore, right=10mm of legs, text width=24mm] (rrf) {Leg weights\\$\downarrow$\\RRF fusion};
\node[icephase, right=10mm of rrf, text width=45mm] (curation) {Sequential curation\\[3pt]
\normalfont bonuses $\to$ session diversity\\$\to$ deduplication\\$\to$ two-phase token budget\\$\to$ strengthen selected memory};
\node[icebox, below=10mm of rrf, text width=65mm] (out) {Selected fragments $\to$ prompt assembler};
\draw[iceflow] (legs) -- (rrf);
\draw[iceflow] (rrf) -- (curation);
\draw[iceflow] (curation.south) |- (out.east);
\end{tikzpicture}%
}
\caption{Retrieval orchestrator. The legs are heterogeneous in retrieval semantics; RRF is what allows them to be combined without score normalisation; the two-phase budget prioritises source-type diversity under budget pressure. Section~\ref{sec:exercised} reports which legs contributed during evaluation.}
\label{fig:orchestrator}
\end{figure}

Prompt assembly, memory slots, and the faulty HTTP wrapper check are documented in Appendix~\ref{app:assembly}; they are not repeated here.

\FloatBarrier
\section{Matched LongMemEval Paired Inference}
\label{app:lme-paired}
All numbers below concern ICE v2. Resampling preserves question identity across arms and, for the degradation contrast, across phases. Intervals are 20,000-resample percentile intervals with seed 20260911. Positive differences favour ICE. Category intervals are exploratory and unadjusted.

\begin{table}[htbp]
\centering\small
\caption{Oracle paired outcomes. CC: both correct; I: ICE only; V: vector only; WW: both wrong. Counts include only paired obtainable verdicts.}
\label{tab:paired-oracle}
\begin{tabular}{@{}lrrrrrl@{}}\toprule
Category & CC & I & V & WW & $\Delta$ (pp) & 95\% CI\\\midrule
Abstention & 17 & 8 & 1 & 4 & +23.3 & $[+6.7,+40.0]$\\
Knowledge update & 37 & 5 & 16 & 14 & -15.3 & $[-27.8,-2.8]$\\
Multi-session & 36 & 0 & 68 & 17 & -56.2 & $[-65.3,-47.1]$\\
Session assistant & 49 & 2 & 4 & 1 & -3.6 & $[-12.5,+5.4]$\\
Session preference & 18 & 3 & 3 & 6 & +0.0 & $[-16.7,+16.7]$\\
Session user & 49 & 1 & 12 & 2 & -17.2 & $[-28.1,-7.8]$\\
Temporal reasoning & 22 & 7 & 32 & 66 & -19.7 & $[-28.3,-10.2]$\\
Overall & 228 & 26 & 136 & 110 & -22.0 & $[-26.6,-17.4]$\\
\bottomrule\end{tabular}
\end{table}

\begin{table}[htbp]
\centering\small
\caption{Full-S paired outcomes. CC: both correct; I: ICE only; V: vector only; WW: both wrong. Counts include only paired obtainable verdicts.}
\label{tab:paired-full}
\begin{tabular}{@{}lrrrrrl@{}}\toprule
Category & CC & I & V & WW & $\Delta$ (pp) & 95\% CI\\\midrule
Abstention & 18 & 7 & 1 & 4 & +20.0 & $[+3.3,+36.7]$\\
Knowledge update & 31 & 6 & 19 & 16 & -18.1 & $[-30.6,-5.6]$\\
Multi-session & 24 & 2 & 66 & 29 & -52.9 & $[-62.8,-43.0]$\\
Session assistant & 41 & 1 & 12 & 2 & -19.6 & $[-32.1,-8.9]$\\
Session preference & 9 & 4 & 6 & 10 & -6.9 & $[-27.6,+13.8]$\\
Session user & 46 & 0 & 14 & 4 & -21.9 & $[-32.8,-12.5]$\\
Temporal reasoning & 22 & 4 & 38 & 63 & -26.8 & $[-35.4,-18.1]$\\
Overall & 191 & 24 & 156 & 128 & -26.5 & $[-31.3,-21.8]$\\
\bottomrule\end{tabular}
\end{table}
The full-S missing vector judgement is in the preference category. Its paired ICE score is 13/29, while the marginal table uses 13/30. The overall phase-degradation contrast uses a common denominator of 499 and therefore differs from subtracting the two displayed marginal accuracy gaps.

\FloatBarrier
\section{Matched LongMemEval Cost Strata}
\label{app:lme-cost-strata}
The following table reports median [25th, 75th percentile] for provider input tokens by category and outcome, plus median selected fragments and generation seconds. C/W are correct/incorrect according to Muse; the single missing verdict is excluded from this stratification. The machine-readable aggregate artifact also includes word-estimated prompt-token distributions and 95th percentiles for every stratum. Retrieval-only context tokens, per-leg candidates, retrieval latency, construction cost per arm, and transient failure rates are unavailable. These tables describe associations, not interventions on context size.
\small
\begin{longtable}{@{}L{1.0cm}L{1.0cm}L{2.6cm}crrrL{3.6cm}@{}}
\caption{ICE v2 public diagnostic: cost conditioned on category and correctness.}\\
\toprule Phase & Arm & Category & C/W & $n$ & Frags & Sec. & Input tokens [IQR]\\\midrule\endfirsthead
\toprule Phase & Arm & Category & C/W & $n$ & Frags & Sec. & Input tokens [IQR]\\\midrule\endhead
Oracle & ICE v2 & Abstention & C & 25 & 3.0 & 3.0 & 1,994 [1,793, 2,055] \\
Oracle & ICE v2 & Abstention & W & 5 & 4.0 & 3.5 & 2,212 [2,085, 2,249] \\
Oracle & ICE v2 & Knowledge update & C & 42 & 4.0 & 2.1 & 2,058 [1,994, 2,108] \\
Oracle & ICE v2 & Knowledge update & W & 30 & 4.0 & 2.7 & 2,077 [2,002, 2,156] \\
Oracle & ICE v2 & Multi-session & C & 36 & 4.0 & 2.8 & 2,090 [1,956, 2,158] \\
Oracle & ICE v2 & Multi-session & W & 85 & 3.0 & 3.0 & 2,081 [2,003, 2,190] \\
Oracle & ICE v2 & Session assistant & C & 51 & 3.0 & 2.3 & 878 [724, 1,023] \\
Oracle & ICE v2 & Session assistant & W & 5 & 3.0 & 3.1 & 991 [729, 1,052] \\
Oracle & ICE v2 & Session preference & C & 21 & 3.0 & 9.6 & 1,959 [1,747, 2,021] \\
Oracle & ICE v2 & Session preference & W & 9 & 3.0 & 6.8 & 1,930 [1,553, 2,010] \\
Oracle & ICE v2 & Session user & C & 50 & 3.0 & 2.0 & 1,710 [1,497, 1,909] \\
Oracle & ICE v2 & Session user & W & 14 & 3.0 & 2.6 & 1,850 [1,522, 2,037] \\
Oracle & ICE v2 & Temporal reasoning & C & 29 & 3.0 & 2.8 & 2,011 [1,931, 2,098] \\
Oracle & ICE v2 & Temporal reasoning & W & 98 & 3.0 & 3.0 & 2,068 [1,934, 2,144] \\
Oracle & ICE v2 & Overall & C & 254 & 3.0 & 2.4 & 1,928 [1,440, 2,066] \\
Oracle & ICE v2 & Overall & W & 246 & 3.0 & 3.0 & 2,065 [1,947, 2,153] \\
Oracle & Vector & Abstention & C & 18 & 11.0 & 2.3 & 5,462 [3,840, 5,817] \\
Oracle & Vector & Abstention & W & 12 & 12.0 & 2.5 & 6,268 [5,060, 6,876] \\
Oracle & Vector & Knowledge update & C & 53 & 12.0 & 1.5 & 5,926 [5,167, 6,502] \\
Oracle & Vector & Knowledge update & W & 19 & 12.0 & 1.5 & 5,658 [5,246, 6,562] \\
Oracle & Vector & Multi-session & C & 104 & 12.0 & 2.2 & 6,544 [5,681, 9,767] \\
Oracle & Vector & Multi-session & W & 17 & 18.0 & 3.5 & 9,324 [6,902, 11,975] \\
Oracle & Vector & Session assistant & C & 53 & 4.0 & 1.6 & 950 [647, 1,384] \\
Oracle & Vector & Session assistant & W & 3 & 6.0 & 2.9 & 1,218 [970, 1,490] \\
Oracle & Vector & Session preference & C & 21 & 7.0 & 7.1 & 3,849 [3,370, 4,245] \\
Oracle & Vector & Session preference & W & 9 & 6.0 & 5.2 & 3,693 [3,521, 4,456] \\
Oracle & Vector & Session user & C & 61 & 6.0 & 1.4 & 3,015 [2,555, 3,441] \\
Oracle & Vector & Session user & W & 3 & 6.0 & 2.0 & 2,970 [2,656, 3,140] \\
Oracle & Vector & Temporal reasoning & C & 54 & 12.0 & 2.0 & 6,496 [5,742, 6,910] \\
Oracle & Vector & Temporal reasoning & W & 73 & 12.0 & 2.3 & 6,388 [4,635, 7,782] \\
Oracle & Vector & Overall & C & 364 & 11.0 & 1.9 & 5,261 [3,014, 6,539] \\
Oracle & Vector & Overall & W & 136 & 12.0 & 2.5 & 6,160 [4,620, 7,712] \\
S & ICE v2 & Abstention & C & 25 & 5.0 & 2.8 & 2,211 [2,150, 2,262] \\
S & ICE v2 & Abstention & W & 5 & 4.0 & 4.3 & 2,259 [2,228, 2,302] \\
S & ICE v2 & Knowledge update & C & 37 & 5.0 & 2.4 & 2,185 [2,139, 2,251] \\
S & ICE v2 & Knowledge update & W & 35 & 5.0 & 2.7 & 2,240 [2,184, 2,292] \\
S & ICE v2 & Multi-session & C & 26 & 6.0 & 2.8 & 2,230 [2,199, 2,278] \\
S & ICE v2 & Multi-session & W & 95 & 5.0 & 3.3 & 2,233 [2,185, 2,298] \\
S & ICE v2 & Session assistant & C & 42 & 6.0 & 2.5 & 2,251 [2,189, 2,292] \\
S & ICE v2 & Session assistant & W & 14 & 6.0 & 3.8 & 2,191 [2,167, 2,242] \\
S & ICE v2 & Session preference & C & 13 & 5.0 & 7.5 & 2,229 [2,217, 2,321] \\
S & ICE v2 & Session preference & W & 17 & 5.0 & 6.8 & 2,223 [2,167, 2,267] \\
S & ICE v2 & Session user & C & 46 & 5.0 & 2.1 & 2,199 [2,148, 2,278] \\
S & ICE v2 & Session user & W & 18 & 5.0 & 3.3 & 2,202 [2,152, 2,288] \\
S & ICE v2 & Temporal reasoning & C & 26 & 5.5 & 3.0 & 2,245 [2,189, 2,297] \\
S & ICE v2 & Temporal reasoning & W & 101 & 5.0 & 3.3 & 2,211 [2,143, 2,264] \\
S & ICE v2 & Overall & C & 215 & 5.0 & 2.6 & 2,224 [2,170, 2,289] \\
S & ICE v2 & Overall & W & 285 & 5.0 & 3.4 & 2,219 [2,167, 2,288] \\
S & Vector & Abstention & C & 19 & 30.0 & 2.6 & 11,391 [10,259, 12,813] \\
S & Vector & Abstention & W & 11 & 30.0 & 3.4 & 11,922 [11,408, 12,202] \\
S & Vector & Knowledge update & C & 50 & 30.0 & 2.0 & 12,026 [10,295, 12,706] \\
S & Vector & Knowledge update & W & 22 & 30.0 & 2.2 & 11,558 [10,738, 12,482] \\
S & Vector & Multi-session & C & 90 & 30.0 & 2.7 & 12,084 [11,153, 13,300] \\
S & Vector & Multi-session & W & 31 & 30.0 & 3.0 & 11,844 [10,791, 13,518] \\
S & Vector & Session assistant & C & 53 & 30.0 & 2.2 & 10,860 [9,118, 11,853] \\
S & Vector & Session assistant & W & 3 & 30.0 & 3.1 & 9,426 [8,277, 11,730] \\
S & Vector & Session preference & C & 15 & 30.0 & 5.5 & 12,494 [11,953, 13,725] \\
S & Vector & Session preference & W & 14 & 30.0 & 6.1 & 12,688 [11,168, 13,912] \\
S & Vector & Session user & C & 60 & 30.0 & 1.9 & 10,802 [9,786, 12,251] \\
S & Vector & Session user & W & 4 & 30.0 & 2.2 & 11,152 [10,630, 11,450] \\
S & Vector & Temporal reasoning & C & 60 & 30.0 & 2.8 & 11,732 [10,954, 13,054] \\
S & Vector & Temporal reasoning & W & 67 & 30.0 & 3.1 & 12,097 [10,625, 13,203] \\
S & Vector & Overall & C & 347 & 30.0 & 2.5 & 11,650 [10,350, 12,800] \\
S & Vector & Overall & W & 152 & 30.0 & 3.1 & 11,854 [10,687, 13,126] \\
\bottomrule\end{longtable}
\normalsize
\FloatBarrier
\section{Historical Local Oracle and Adapter Invalidation}
\label{app:lme-history}
Before the matched cloud study, the corrected v2 adapter used \texttt{gemma4:26b-a4b-it-q4\_K\_M} for both answer arms and local \texttt{gemma4:12b} judging. ICE obtained 264/478 (55.2\%) and vector 388/484 (80.2\%) on the oracle; 22 and 16 judge mutes gave all-500 bounds of 52.8--57.2\% and 77.6--80.8\%. That study stopped before full-S. Its scores are historical diagnostics, not part of the matched cloud comparison, and the later completed cloud phases supersede the stopped-phase statement.

The earlier flattened-session adapter is invalid. A trace showed 45 lexical plus 100 vector candidates, 111 unique after fusion, then only three after diversification because a UUID object did not equal a string identifier. Its query placement also failed to preserve supplied session boundaries. These results remain excluded. The corrected session adapter separates history sessions and queries from an empty conversation; it does not inflate the fresh-session retrieval budget from the total haystack length.

\FloatBarrier
\section{LSREP Sensitivity and Reproduction}
\label{app:clustered}
The historical record-level bootstrap treats repeated checkpoint observations as separate sampling units. A sensitivity analysis resamples 219 distinct probe clusters, retaining all observations within each selected probe and the ICE/vector pair. It targets the same observation-weighted score difference rather than giving every distinct probe equal weight. With 20,000 resamples and seed 20260911, all-data $\Delta=+0.396$, CI $[+0.192,+0.618]$; ordinary-density $\Delta=+0.002$, CI $[-0.148,+0.158]$ (182 clusters); density-only $\Delta=+3.097$, CI $[+2.721,+3.470]$ (37 clusters). Thus the principal qualitative findings survive while the uncertainty is wider. Conversations and users are not independently resampled; no population claim follows.

The public analysis entry points are \path{experiments/lme/analyze_matched.py} and \path{experiments/mature/clustered_sensitivity.py}. The former reads local arm verdicts and cost metadata and exports only aggregate cells and distributions; the latter reuses the archived manual-merge and score-imputation semantics. No models are rerun. Both must be run from the repository root through \texttt{uv run python}. The aggregate files contain no question text, answers, judgements, or personal identifiers.

\subsection{Scoring-Source, Missing-Score, and Ordinal Sensitivity}
\label{app:scoring}
Table~\ref{tab:scoring-sensitivity} crosses two choices: retaining versus removing the 72 manual replacements, and using the archived fallback chain versus requiring explicit scores in both generalist arms. All intervals resample entire probe trajectories with pairs preserved (20,000 resamples, seed 20260911). Complete-case selection is informative: a failed vector answer often has no explicit score. The automatic-only complete-case density subset contains just 13 of 154 observations, and therefore says little about reliability over the original workload. No missing-at-random assumption is made.

\begin{table}[htbp]
\centering\small
\caption{ICE v2 minus vector generalist under alternative scoring policies. ``Archived'' includes failure assignment and fallback; ``complete'' requires two explicit scores. These are sensitivity analyses of recorded outcomes, not new model runs.}
\label{tab:scoring-sensitivity}
\input{generated/scoring_sensitivity.tex}
\end{table}

The merged generalist vector scores comprise 1,079 explicit labels, 130 failed-answer assignments to 1, and two sibling-routing substitutions. Without manual replacements the counts are 1,067, 141, and three. ICE generalist has 1,211 explicit scores under both policies. Across the other two routing arms, vector-MoE has 34 failed-answer assignments and ICE-MoE one sibling substitution. Neither policy uses the rounded-record-mean or default-3 fallback. The analysis reproduces every merged score from the archived aggregation before estimating sensitivity.

As an ordinal check, we use the direction-only principle of paired sign comparisons~\citep{nistSignTest}, reporting paired win/tie/loss counts and the net superiority proportion $\Pr(S_{\mathrm{ICE}}>S_{\mathrm{vector}})-\Pr(S_{\mathrm{ICE}}<S_{\mathrm{vector}})$. We use probe-cluster bootstrap intervals rather than an independent-observation binomial sign test. This compares ordering alone; it does not claim that a change from 1 to 2 equals a change from 4 to 5. In all data, the merged scores give 355 ICE-higher observations, 220 vector-higher, and 636 ties: net $+11.1$ points (cluster CI $[+3.2,+19.5]$). Ordinary-density net superiority is $+0.1$ ($[-7.6,+8.0]$), and density-only is $+87.0$ ($[+77.5,+95.3]$). Failure assignment remains part of this ordinal reliability comparison. These results supplement the historical mean rubric scores; SPF and TUR remain descriptive engineering summaries, not validated substitutes for answer quality.

The reproducible entry point is \path{experiments/mature/scoring_sensitivity.py}; \path{experiments/paper/generate_analysis_tables.py} emits the aggregate TeX table and shared ablation macros. The ablation report now reuses each paired contrast across its step, cumulative, and headline views, eliminating small Monte Carlo discrepancies caused by resampling the same contrast repeatedly.

\end{document}

%% file: generated/ablation_macros.tex
\newcommand{\LexicalCI}{[-1.14,-0.36]}

\newcommand{\FusionCI}{[+0.39,+1.24]}
\newcommand{\FusionBareCI}{[-0.18,+0.34]}
\newcommand{\FullBareCI}{[-0.24,+0.42]}
\newcommand{\FullVectorCI}{[-0.38,+0.25]}

%% file: generated/scoring_sensitivity.tex
\begin{tabular}{llrrr}
\toprule
Scoring & Regime & $n$ & $\Delta$ & Probe-cluster 95\% CI \\
\midrule
Merged, archived & All & 1,211 & $+0.396$ & $[+0.192,+0.618]$ \\
Merged, archived & Ordinary & 1,057 & $+0.002$ & $[-0.148,+0.158]$ \\
Merged, archived & Density & 154 & $+3.097$ & $[+2.721,+3.470]$ \\
Automated, archived & All & 1,211 & $+0.363$ & $[+0.162,+0.582]$ \\
Automated, archived & Ordinary & 1,057 & $-0.020$ & $[-0.169,+0.136]$ \\
Automated, archived & Density & 154 & $+2.994$ & $[+2.606,+3.359]$ \\
Merged, complete & All & 1,079 & $+0.046$ & $[-0.101,+0.199]$ \\
Merged, complete & Ordinary & 1,055 & $+0.001$ & $[-0.148,+0.157]$ \\
Merged, complete & Density & 24 & $+2.042$ & $[+1.077,+3.294]$ \\
Automated, complete & All & 1,067 & $-0.021$ & $[-0.168,+0.128]$ \\
Automated, complete & Ordinary & 1,054 & $-0.022$ & $[-0.170,+0.131]$ \\
Automated, complete & Density & 13 & $+0.077$ & $[-0.714,+0.714]$ \\
\bottomrule
\end{tabular}